\documentclass[letterpaper,twocolumn,10pt]{article}
\usepackage{arxiv}
\usepackage{graphicx}
\usepackage{url}
\usepackage{multirow}
\usepackage[inline]{enumitem}

\usepackage{booktabs}
\usepackage{tabularx}
\usepackage{adjustbox}
\usepackage{booktabs,adjustbox,array,makecell}
\usepackage{xcolor}

\usepackage{tikz}
\usepackage[most]{tcolorbox}
\usepackage{pgf-pie}
\usepackage{tikz}
\usetikzlibrary{calc}
\usepackage{xparse}
\usepackage{ifthen}
\usetikzlibrary{arrows.meta,positioning,fit,calc,shapes.geometric}
\usepackage{xspace}

\usepackage{cite}
\usepackage{amsmath,amssymb,amsfonts}
\usepackage{textcomp}
\usepackage{algorithm}
\usepackage{algpseudocode}

\newtcolorbox{rqBox}{
    sharpish corners, 
    boxrule = 0pt,
    toprule = 4.5pt, 
    enhanced,
    fuzzy shadow = {0pt}{-2pt}{-0.5pt}{0.5pt}{black!35}
}

\newcommand{\replicant}{\textsc{Replicant}\xspace}
\newcommand{\replicantWB}{\textsc{Replicant$_{WB}$}\xspace}
\newcommand{\apgWB}{APG$_{WB}$\xspace}
\newcommand{\atreplicant}{\textsc{AT-Replicant}\xspace}
\newcommand{\apgat}{AT-APG\xspace}

\definecolor{gaincol}{RGB}{20,130,20}
\definecolor{losscol}{RGB}{200,30,30}
\definecolor{neucol}{RGB}{120,120,120}

\usepackage[nohyperlinks]{acronym}

  \acrodef{DRL}{Deep Reinforcement Learning}
  \acrodef{RL}{Reinforcement Learning}
  \acrodef{MDP}[MDP]{Markov Decision Process}
  \acrodefplural{MDP}[MDPs]{Markov Decision Processes}
  \acrodef{POMDP}[POMDP]{Partially Observable \ac{MDP}}
  \acrodefplural{POMDP}[POMDPs]{Partially Observable \acp{MDP}}

  \acrodef{ML}{Machine Learning}

  \acrodef{AML}{Adversarial Machine Learning}
  \acrodef{AE}[AE]{Adversarial Sample}
  \acrodefplural{AE}[AEs]{Adversarial Samples}
  \acrodef{AT}{Adversarial Training}

  \acrodef{AL}{Active Learning}
  \acrodef{RAL}{Robust \ac{AL}}

  \acrodef{PTE}{Policy Transfer Evasion}
  \acrodef{STE}{Sample Transfer Evasion}

  \acrodef{ASR}{Attack Success Rate}

\title{\Large \bf \textsc{Replicant}: Learning Policies for Evading and Hardening Malware Detectors}

\author{
    {\rm  
        {\bfseries Shae McFadden}\textsuperscript{\rm$\dagger$\rm$\ddagger$\rm$\mathsection$}, 
        {\bfseries Ilias Tsingenopoulos}\textsuperscript{\rm$\P$},
        {\bfseries Mario D'Onghia}\textsuperscript{\rm$\mathsection$}, 
        {\bfseries Alexander Herzog}\textsuperscript{\rm$\mathsection$\rm$\parallel$},
    } \\ {\rm 
        {\bfseries Myles Foley}\textsuperscript{\rm$\ddagger$\rm*},
        {\bfseries Chris Hicks}\textsuperscript{\rm$\ddagger$}, 
        {\bfseries Lorenzo Cavallaro}\textsuperscript{\rm$\mathsection$}
        {\bfseries Fabio Pierazzi}\textsuperscript{\rm$\mathsection$}
    } \\ 
        \textsuperscript{\rm  $\dagger$}\textit{King's College London}, 
        \textsuperscript{\rm $\ddagger$}\textit{The Alan Turing Institute}, 
        \textsuperscript{\rm  $\mathsection$}\textit{University College London}, 
    \\ 
        \textsuperscript{\rm  $\P$}\textit{KU Leuven},
        \textsuperscript{\rm  $\parallel$}\textit{Core64}, 
        \textsuperscript{\rm  *}\textit{Devotion AI Labs}
}

\begin{document}

\maketitle

\begin{abstract}
    To determine the real-world effectiveness of \textit{machine learning} based malware detection, it is vital to evaluate its robustness against highly capable adversaries.
    However, state-of-the-art attacks do not effectively model realistic adversaries, as they often assume access to privileged information such as the training data, feature space, or confidence scores of the target.
    In this work, we present \replicant, a deep reinforcement learning framework that \textit{learns the realistic task} of evasion under \textit{a strict label-only black-box threat model}. 
    \replicant learns a reusable policy on \textit{how} to modify a malware sample and \textit{when} to query the target, which transfers across samples, detectors, and feature spaces.  
    Across seven Android malware detectors and three feature spaces, \replicant is the strongest and most query-efficient approach achieving a mean attack success rate of $78.8\%$, a relative improvement of $20.9\%$--$39.2\%$ over the state-of-the-art. 
    Furthermore, when used for adversarial training, \replicant also outperforms the state-of-the art by producing detectors with more generalizable robustness. 
    With \replicant we demonstrate that learning the task of evasion not only results in stronger attack performance but, crucially, provides a better signal for hardening malware detectors.
\end{abstract}

\section{Introduction} 
\label{sec:introduction}

    \ac{ML} classifiers have become a key line of defence against malware, with extensive research focusing on improving their performance and robustness to drift~\cite{Drebin,zhang2020enhancing,ramda,MaMaDroid,mcfadden2026drmd,chen2023continuous,Transcend,Transcendent,zheng2026tif,li2025revisit} and Android being an ideal target for malware as it is the most widely used mobile operating system globally~\cite{AndroidStatistica}. 
    However, the field of \ac{AML} has repeatedly demonstrated the brittleness of \ac{ML}-based systems under targeted modifications~\cite{goodfellow2014explaining,carlini2017towards,biggio2013evasion,WildPatterns}.
    Although well established, this vulnerability is often underpinned with strong assumptions, such as access to confidence scores; therefore, the question most relevant for risk assessment is not \textit{whether} these detectors can be evaded, but how \textit{general} and \textit{efficient} an attack can be under realistic constraints.

    Most evasion research operates in the \textit{feature space}, directly modifying the input that is fed into \ac{ML} systems~\cite{goodfellow2014explaining,carlini2017towards,WildPatterns}. 
    Although sufficient to evade systems such as image detection, these approaches are not applicable to crafting evasive malware due to the \textit{inverse mapping problem}: a change to the feature vector does not necessarily correspond to any realizable and function-preserving change in the application~\cite{pierazzi2020intriguing,cortellazzi2025intriguing,bostani2024evadedroid,he2023efficient,song2022mab,scano2026droidbreaker,quiring2019misleading}.   
    Alternatively, \textit{problem-space} attacks avoid this problem by modifying the application directly while enforcing syntactical and semantic constraints; therefore, every adversarial example is a valid program by construction. 

    Despite the shift towards more realistic problem-space evasion, the Android evasion literature is still \textit{fragmented} with three major limitations. 
    First, existing works optimize individual samples against a single target classifier and feature space, relying on confidence scores and search heuristics or greedy selection~\cite{pierazzi2020intriguing,bostani2024evadedroid,he2023efficient}. 
    As a result, these attacks optimize on a per-sample basis without retaining knowledge from past evasions, capping their query efficiency and limiting their transferability to new targets.
    Second, existing literature often conflates two separate components of problem-space attacks: (i) the \textit{capability set}, which defines modifications available to the attacker that can be made while preserving problem-space constraints; and, (ii) the \textit{strategy}, which selects from the capability set in order to modify a sample to evade a target.
    Therefore, when both components are proposed together and not evaluated in isolation, it is unclear if the new attack has discovered a better strategy for evasion, or if a potentially sub-optimal strategy is being carried by a stronger capability set.
    This ambiguity is a form of the gain attribution pitfall raised by~\cite{mcfadden2026sok}. 
    Finally, these approaches are evaluated only on a single temporal snapshot, ignoring the impact that temporal drift and targets that employ \ac{AL}~\cite{settles2009active} have on the evasion of malware samples. 
    We therefore lack a comprehensive understanding of how evasion strategies generalize and transfer across \textit{classifiers}, \textit{feature spaces}, and \textit{time}.

    To address these limitations, we propose \replicant: a unified \ac{DRL} framework that \textit{learns} problem-space evasion policies. 
    Evasion here is formulated as a sequential decision-making problem (specifically, a Markov Decision Process) where an agent observes the current representation of a modified malware sample and chooses either: (i) to submit the sample to the target; or, (ii) to modify the sample and select which capability to apply.
    \replicant is deliberately agnostic to \textit{what} the problem-space capabilities are: it learns effective evasion policies given \textit{any} capability set and a representation that captures how the modifications affect the sample.
    For Android malware detection, one of the most common capabilities is the harvesting and transplantation of self-contained code fragments from benign applications~\cite{pierazzi2020intriguing,cortellazzi2025intriguing,bostani2024evadedroid,he2023efficient}.
    Concretely, we instantiate our capability set using gadget transplantation as proposed in~\cite{pierazzi2020intriguing,cortellazzi2025intriguing}.
   
    In this work we evaluate \replicant under five key settings.
    First, we establish \replicant as a state-of-the-art black-box attack, comparing it to prior attacks~\cite{pierazzi2020intriguing,bostani2024evadedroid,he2023efficient,zhan2023psp,quertier2022merlin,song2022mab,rigaki2023power,labaca2021aimed}, in the largest Android malware evasion study to date.
    Second, we demonstrate the advantage of learned policy-transfer over sample-transfer across all $1{,}764$ surrogate and target combinations.
    Third, we investigate the defensive utility of the \replicant attack through the use of a white-box variant (\replicantWB) to produce robust targets through \ac{AT}, \atreplicant, comparing it with the white-box attack APG (\apgWB /\apgat)~\cite{pierazzi2020intriguing}.
    Fourth, we assess how well this robustness from \ac{AT} withstands an evolving attacker capability set.
    Lastly, we explore the interplay between \ac{AL} and \ac{AT} when evaluating both clean performance and adversarial robustness over time.
    In summary, we make the following contributions:
    \begin{itemize}
        \item \textbf{\replicant Framework.} 
        We propose \replicant, a general \ac{DRL}-based framework that learns problem-space evasion over a capability set as a \textit{hierarchical policy}, improving \ac{ASR} by $20.9\%$--$39.2\%$ compared to state-of-the-art Android attacks. 
        
        \item \textbf{Policy Transfer.} 
        We demonstrate that a policy trained against a surrogate transfers more effectively than the samples that evade the same surrogate, yielding an $82.0\%$ relative increase in \ac{ASR}.

        \item \textbf{\ac{AT} \& Capability Drift.} 
        We show that \atreplicant reduces attacker success below $17\%$ for \textit{both} \replicantWB and a \apgWB, whereas \apgat leaves the model over five times more vulnerable to \replicantWB.
        We also show that the robustness gained from \ac{AT} diminishes as the attacker capabilities expand from the capability set used by the defender.

        \item \textbf{Robustness under Drift.}
        We establish that applying \ac{AL} directly to \ac{AT} hardened classifiers recovers drift-induced performance but degrades robustness, whereas our \ac{RAL} retains the performance-over-time of \ac{AL} and the robustness of \ac{AT}.
        
        \item \textbf{Comprehensive Study.} 
        We conduct the largest evaluation of problem-space Android malware evasion to date, comprising $379{,}680$ attack evaluations, isolating the gains of evasion strategies on attacker success.
    \end{itemize}

\section{Background} \label{sec:background}

\subsection{Android Malware Detection} \label{sec:AMD}

    We evaluate this work in the Android malware domain due to the extensive work eliminating experimental bias in evaluations~\cite{Tesseract} and dataset composition~\cite{chow2025breaking} with AndroZoo~\cite{androzoo} supplying the large-scale, representative, and timestamped corpus on which these bias mitigations depend on.

    \noindent\textbf{Representations.} 
    The dominant pipeline in Android malware detection consists of extracting \textit{static} features from the manifest, alongside the bytecode of the application, and encoding each sample as a high-dimensional binary vector. 
    The three feature spaces used in this work are: 
    \textit{Drebin}~\cite{Drebin}, the de facto standard in the domain, which spans permissions, API calls, intents, hardware components, and network addresses; 
    \textit{APIGraph}~\cite{zhang2020enhancing}, which enriches the \textit{Drebin} feature space with clustering-based representations. \textit{APIGraph} maps functionally similar APIs to the same primitive, which reduces temporal  drift due to software-evolution. 
    Lastly, \textit{RAMDA}~\cite{ramda}, which, as a subset of the \textit{Drebin} feature space, covers API calls, intents, and permissions. 
    The variable size and overlap between these three feature spaces allows us to evaluate scenarios in which the attacker has varying degrees of knowledge of the feature space the target detector is trained on.

    \noindent\textbf{Detection.} 
    Based on these features, the community has developed a diverse set of detection methodologies: 
    linear models ranging from the original Drebin SVM~\cite{Drebin} to the robust variant SecSVM~\cite{demontis2017yes}; ensemble methods such as Random Forests (RF)~\cite{MaMaDroid,wu2025malscan}; and deep learning approaches including DeepDrebin~\cite{deepDrebin}, the robust android malware detection algorithm (RAMDA) classifier~\cite{ramda}, the hierarchical contrastive classifier (HCC)~\cite{chen2023continuous}, and the deep reinforcement malware detector (DRMD)~\cite{mcfadden2026drmd}. 
     This spectrum allows us to assess the general performance and transferability of attacks between diverse models and detection methodologies.

    \noindent\textbf{Temporal Drift.} 
    Finally, a major challenge in malware detection is \textit{temporal drift}. We use temporal drift as an umbrella term for covariate shift, prior probability shift, and concept shift~\cite{moreno}.
    Since both goodware and malware evolve over time, the test distribution diverges from the training distribution, degrading performance-over-time~\cite{Tesseract,kan2024tesseract,mcfadden2023poster,mcfadden2024impact,chow2025breaking}. 
    Therefore, training without temporally consistent splits or testing on unrealistic class-ratios can substantially inflate performance~\cite{Tesseract,kan2024tesseract}. 
    We follow the recommendations of \textit{Tesseract}~\cite{Tesseract} and the guidelines of~\cite{DosDonts}, adopting time-aware splits with a realistic malware distribution ($\approx10\%$) and an explicit temporal evaluation. 
    Critically, as malware authors continuously adapt to avoid detection and improve functionality, drift in this domain is, at least in part, intrinsically intertwined with the objective of evasion.

\subsection{Evasion Attacks}

    Evasion attacks perturb the inputs of an \ac{ML} system at test-time in order to cause misclassification. 
    Specifically, in the context of malware detection, evasion attacks modify malware samples to be classified as goodware while retaining their original behavior. 
    In practice, access to the actual target model is limited; thus, the two predominant strategies of transferability and querying attacks were developed to make attacks practical. 
    \textit{Transferability} attacks leverage a surrogate model to optimize an \ac{AE} and rely on decision boundary similarity to evade the target~\cite{AttackTransfer}. 
    Alternatively, \textit{Querying} attacks interact with the target directly, submitting successive versions of an \ac{AE} and using the feedback received to guide evasion while minimizing queries~\cite{brendel2017decision,chen2020hopskipjumpattack,debenedetti2024evading}. 
    \replicant combines both of these methodologies by learning an evasion policy on a surrogate (transferability) and then uses that policy to choose \textit{what} to modify and \textit{when} to submit new malware samples to the target (querying).
    
    \noindent\textbf{Inverse Mapping Problem.} 
    Feature-space attacks perturb the feature vector directly, similarly to evasion attacks in computer vision, where pixels are modified via gradient-based perturbations~\cite{ding2020adversarial,goodfellow2014explaining,kurakin2018adversarial,dong2018boosting,madry2017towards,moosavi2016deepfool,carlini2017towards,shereen2025one}. 
    In the context of the predominantly binary feature spaces of Android malware detection, adding a feature ($0\to1$) requires concretely introducing the corresponding code (such as API or permission), while removing a feature ($1\to0$) may require removing or breaking functionality that is central to the malware. 
    As a result, when the representation reflects real programs, the feature extraction process is often lossy, giving rise to the \textit{inverse mapping problem}~\cite{pierazzi2020intriguing,cortellazzi2025intriguing} as the feature mapping from the problem space is neither invertible nor differentiable.

    \noindent\textbf{Problem-Space Evasion.} 
    Problem-space attacks circumvent the inverse-mapping problem by grounding modifications in the program itself, with each transformation mapping actual code to a known feature space effect~\cite{pierazzi2020intriguing,cortellazzi2025intriguing}. 
    Using the formalization by~\cite{pierazzi2020intriguing}, problem-space attacks must meet the following criteria: 
    (i) \textit{available transformations (realizability)}, using only modifications an attacker could perform in practice; 
    (ii) \textit{preserved semantics}, the transformations performed must preserve the original malicious behavior of the sample; 
    (iii) \textit{plausibility}, the sample should appear realistic upon manual inspection; 
    (iv) \textit{robustness to preprocessing}, the modifications should survive a realistic pre-processing pipeline. 
    The most established way to satisfy these constraints for problem-space Android malware attacks is \textit{gadget extraction and transplantation}, where self-contained code fragments alongside their known features are harvested from benign applications and transplanted into malware~\cite{pierazzi2020intriguing,bostani2024evadedroid,cortellazzi2025intriguing,he2023efficient,jigsaw,bostani2024effectiveness}. 
    As a result of gadgets being harvested from real applications, they are realizable and plausible by construction. 
    Furthermore, since gadget transplantation only \textit{adds} benign code, and thus features ($0\to1$), it preserves the original behavior (semantics) of the malware. 
    Therefore, given a set of gadgets, problem-space evasion is then reduced to finding effective strategies over that capability set, which we pose as a general \ac{DRL} problem rather than per-sample optimization.
    
    \noindent\textbf{Adversarial Training.}
    Despite a long history of proposed defenses against evasion attacks, adversarial training remains one of the few defenses that has consistently demonstrated empirical robustness~\cite{madry2017towards,10179316}.
    Evasion attacks highlight the weakness in a models decision-making; by retraining against these \acp{AE} the model can become \emph{hardened} or more robust against such attacks.
    Given training data $D=\{(x_i,y_i)\}_{i=1}^{n}$, where $x_i$ denotes a malware sample represented in an arbitrary feature space $\mathcal{X}$ and $y_i \in \{1,..., Y\}$ is its class label, adversarial training solves the following min-max optimization problem:

    \begin{equation}
        \min_{\phi}\;
        \mathbb{E}_{(x_i,y_i)\sim D}
        \left[
        \max_{x_i' \in \Delta(x_i)}
        \mathcal{L}\!\left(\mathrm{clf}_{\phi}(x_i'),\, y_i\right)
        \right],
    \label{eqn:adv_train}
    \end{equation}
    
    \noindent where $\mathrm{clf}_{\phi}:\mathcal{X}\rightarrow\mathbb{R}^{Y}$ denotes the classifier parameterized by $\phi$, $\mathcal{L}$ is the training loss, and $\Delta(x_i)$ defines the set of admissible perturbations for sample $x_i$. In the malware domain, $\Delta(x_i)$ typically consists of functionality-preserving modifications that satisfy application-specific constraints.
    The inner maximization seeks a perturbation that maximizes the classifier loss within the admissible perturbation set, while the outer minimization updates the model parameters to minimize the expected adversarial loss, typically using stochastic gradient descent.

\subsection{Deep Reinforcement Learning}
    Fundamentally, reinforcement learning (RL) problems learn optimal policies through environmental interactions formalized as Markov Decision Processes (MDPs). 
    MDPs are defined by the tuple $(\mathcal{S}, \mathcal{A}, \mathcal{R}, \mathcal{T}, \gamma)$, consisting of: a state space $\mathcal{S}$, action space $\mathcal{A}$, reward function $\mathcal{R}: \mathcal{S} \times \mathcal{A} \rightarrow \mathbb{R}$, transition function $\mathcal{T}: \mathcal{S} \times \mathcal{A} \rightarrow \Delta(\mathcal{S})$ (where $\Delta(\mathcal{S})$ denotes probability distributions over states), and discount factor $\gamma \in [0,1]$~\cite{kaelbling1996reinforcement, arulkumaran2017deep, sutton2018reinforcement}. 
    An RL `agent' observes a state $s_t$, selects an action $a_t$ according to the policy $\pi$, then receives a reward $r_t$, and transitions to the next state $s_{t+1}$. 
    The agent then learns a policy that maximizes the expected cumulative discounted reward: $\mathbb{E}[\sum_{t=0}^{\infty} \gamma\ ^t r_t]$.

    \noindent\textbf{Proximal Policy Optimization (PPO).} 
    PPO~\cite{schulman2017proximal}, one of the most widely used \ac{DRL} approaches, is an actor-critic method which parametrizes the policy $\pi$ (actor) alongside the state-value function $V(s_t)$ (critic) with neural networks and optimizes a combined objective:
    \begin{equation} \label{eq:ppo-loss}
      \mathcal{L}^{\mathrm{PPO}}(\theta) \;=\;
      -\,\mathcal{L}^{\mathrm{P}}(\theta)
      \;+\; c_{v}\,\mathcal{L}^{V}(\theta)
      \;-\; c_{e}\,\mathcal{L}^{E}(\theta).
      \end{equation}
    where $\mathcal{L}^{\mathrm{P}}$ is the clipped surrogate objective, $c_{v}$ is the value coefficient, $\mathcal{L}^{V}$ is the clipped value loss, $c_e$ is the entropy coefficient, and $\mathcal{L}^{E}$ is the entropy bonus. 
    Further breakdown of PPO can be found in Appendix~\ref{app:ppo}.

    \noindent\textbf{Action Masking.} 
    In an environment only a subset of the action space may be applicable in a given state, as in problem-space evasion where capabilities may conflict with one another or add nothing new to the current sample.
    Action masking handles this by masking out the logits of inapplicable actions before the softmax, so that they receive near zero probability and are never sampled.
    Therefore, this improves the learning efficiency of \ac{DRL} in tasks with large and conditionally applicable action spaces, while maintaining valid policy-gradient updates~\cite{huang2020closer}.

    \noindent\textbf{Hierarchical Policies.} 
    Hierarchical \ac{DRL}~\cite{kulkarni2016hierarchical} splits a policy across different levels of abstraction, where a higher-level policy decides what to do next and a lower-level policy decides how to do it. 
    Hierarchical \ac{DRL} has been shown to improve the performance and tractability of learning in security problems~\cite{foleyCAGEI22,foleyCAGEII22,foley2022haxss,faillon2024better}.

    \noindent\textbf{DRL for Cybersecurity.} 
    DRL having been used to achieve impressive performance in numerous cybersecurity tasks~\cite{al2023sqirl,foley2022haxss,mcfadden2024wendigo,hicks2026building,mcfadden2026drmd,tsingenopoulos2022captcha,foleyCAGEII22,terranova2024leveraging,faillon2024better,lee2022link}.
    However, recent work has raised the common pitfalls of \ac{DRL} for security research~\cite{mcfadden2026sok}; we therefore follow these best practices to ensure effective demonstration of our modeling, training, evaluation, and deployment assumptions.

\section{\replicant Framework} \label{sec:framework}
    This section describes the \replicant framework, which formulates problem-space evasion as an MDP and uses hierarchical PPO to learn an attack policy.
    Our formulation treats the set of available problem-space modifications as an abstract \textit{capability set}, where each element has an observable effect on the feature representation used by the attacker.
    Throughout our experiments, we use Android gadget transplantation following~\cite{pierazzi2020intriguing,cortellazzi2025intriguing,jigsaw,bostani2024effectiveness} as the concrete instantiation of the capability set.
    Notably, our MDP formulation is general and independent of the capability set or the underlying malware detector, thus applicable to any domain with problem-space transformations that have observable effects to the attacker; this general property of RL-based attacks has also been formalized in \cite{tsingenopoulos2024train}.
    Subsection~\ref{sec:threat-model} explicitly describes our threat model, Subsection~\ref{sec:replicant-mdp} defines our environment (MDP), and Subsection~\ref{sec:replicant-agent} presents our agent.

    \subsection{Threat Model}\label{sec:threat-model}

        \noindent\textbf{Attacker Goal.} 
        The goal of the attacker is to violate the \textit{integrity} of the system by modifying malware to be classified as goodware while preserving its malicious behavior. 
        Furthermore, in the process of achieving this goal, the attack aims to minimize the number of both malware modifications and target queries.

        \noindent\textbf{Attacker Capabilities.} 
        The attacker has a \textit{capability set}: a finite collection of modifications that can be applied to a malware sample while preserving its behavior, each with an observable effect on the feature representation used by the attacker.
        Throughout this paper we instantiate the capability set with \textit{gadgets}, self-contained code fragments harvested from benign applications and transplanted into the malware sample, following~\cite{pierazzi2020intriguing,cortellazzi2025intriguing,jigsaw,bostani2024effectiveness}.

        \noindent\textbf{Attacker Knowledge.} 
        This paper considers two levels of attacker knowledge: black-box used in Sections~\ref{sec:eval-query}--\ref{sec:eval-transfer} and white-box used in Sections~\ref{sec:eval-at}--\ref{sec:eval-temporal}.
        In the black-box setting, the attacker has no explicit knowledge of the target classifier $\mathrm{clf}_{target}$ and receives only its predicted label (i.e., either benign or malicious), rather than a confidence score. 
        To learn an evasion policy, the attacker instead uses a \textit{surrogate} classifier $\mathrm{clf}_{surrogate}$ created by themselves or others, which is trained on distinctly different data from the target.
        Therefore, the surrogate may also differ from the target in both classifier architecture or feature space, which we evaluate in Sections~\ref{sec:eval-query}--\ref{sec:eval-transfer}.
        In the white-box setting, the attacker has unrestricted access to the target and thus may use its confidence scores and feature-space representation.
        We adopt the white-box threat model for the \ac{AT} experiments in Sections~\ref{sec:eval-at}--\ref{sec:eval-temporal} for the following reasons.
        First, a defender hardening their own model has complete access to it and thus would use the strongest realistic attacker available to them. 
        Second, isolating the robustness gained from \ac{AT} is best measured using white-box attacks as adding black-box constraints conflates robustness with the difficulty of finding an evasive strategy without access to the target.

        \noindent\textbf{Attacker Parity.}
        To fairly compare attacks, we evaluate prior attack baselines with the same capability set and threat model as \replicant.
        In the black-box attack settings, all privileged information is drawn from the surrogate, and the target only provides predicted labels when queried.
        In the white-box settings used for \ac{AT}, the attackers have direct access to the confidence scores of, and no limits on the interactions with, the target.
        As mentioned in Section~\ref{sec:introduction}, by eliminating the confounding factor of differing capability sets, this allows our evaluation to isolate the performance differences between attack strategies.

    \subsection{\replicant MDP}\label{sec:replicant-mdp}

        \noindent\textbf{State Space.} 
        The state $s_t\in\mathcal{X}$ is the current representation of the (partially modified) sample after $t$ steps, in the feature space $\mathcal{X}$ used by the attacker, with $s_0$ being the original representation of the malware.

        \noindent\textbf{Action Space.} 
        The hierarchical action space of \replicant is split across two separate policies: the higher-level \textit{query} policy and the lower-level \textit{modify} policy.
        Based on $s_t$, \replicant first selects a \textit{query} action, $q_t\in\{$\textsc{Submit},\textsc{Modify}$\}$. If {$q_t$=\textsc{Submit}} then the classifier is queried, $\mathrm{clf}(s_t)\in\{0,1\}$, and no capabilities are applied. 
        If {$q_t$=\textsc{Modify}}, then the modify policy selects a capability to apply to the malware sample and does not query the classifier. 
        The \textit{modify} action $m_t\in\{1,\dots,N\}$ selects a capability $c_{m_t}: \mathcal{X}\rightarrow\mathcal{X}$ from the capability set $C=\{c_1,\dots,c_N\}$. 
        \textit{Action masking} removes, prior to selection, any modify action whose capability either: 
        (i)~adds no new features (a \textit{no-op}), or
        (ii)~would \textit{conflict} in the problem space, covered in Appendix~\ref{app:hyperparams}.
        Furthermore, masking is also applied to the query policy: 
        (i)~\textsc{Modify} is masked every $p$ steps during training ($p=10$), forcing periodic submits, to ensure the agent receives feedback from the classifier throughout training episodes; and 
        (ii)~\textsc{Submit} is masked until at least one modification has occurred since the last query.

        \noindent\textbf{Transition Function.} 
        The state transitions are driven by the modify actions ($m_t$), as each applied capability modifies the underlying sample and, therefore, updates its feature representation.
        Since each capability corresponds to a concrete problem-space modification whose effect on the feature space is known in advance, we can model and evaluate the MDP in the feature space while maintaining the problem-space realizability of every transition.
        The episode terminates when the step count reaches the horizon $H$ (max steps per episode) or a query evades the classifier. Therefore, the full transition function is as follows:
        \begin{equation} \label{eq:termination}
         \resizebox{0.4\textwidth}{!}{$
            T(s_t,q_t,m_t) = \begin{cases}
                terminal & \text{if }q_t=\textsc{Submit} \land \mathrm{clf}(s_t)=0, \\
                terminal & \text{if }t = H, \\
                c_{m_t}(s_t) & \text{if }q_t=\textsc{Modify}, \\
                s_t & \text{else. }
            \end{cases}
            $}
        \end{equation}
    
        \noindent\textbf{Reward Function.} 
        The reward function is as follows: 
        \begin{equation} \label{eq:reward}
        \resizebox{0.43\textwidth}{!}{$
            R(s_t,q_t, m_t)=\begin{cases}
            R_{\text{success}} & \text{if }q_t=\textsc{Submit} \land \mathrm{clf}(s_t)=0, \\
            R_{\text{detected}} & \text{if }q_t=\textsc{Submit} \land \mathrm{clf}(s_t)=1, \\
            R_{\text{modify}}(s_t,c_{m_t}) & \text{if }q_t=\textsc{Modify}.
        \end{cases}
        $}
        \end{equation}
        where $R_{\text{success}}\gg0$ rewards successful evasion, $R_{\text{detected}}\!\le\!0$ penalizes detection, and $R_{\text{modify}}(s_t,c_{m_t})\!<\!0$ penalizes each capability applied. 
        The relative magnitudes of these three terms determine the trade-offs between modification and query efficiency learned by the agent. 
        In this paper, we use:
        \begin{equation} \label{eq:reward-values}
        R_{\text{success}} = H, \;\; 
        R_{\text{detected}} = -1, \;\;  
        R_{\text{modify}}(s_t,c_{m_t}) = -1.
        \end{equation}
        These values were chosen to: 
        (i) ensure every successful episode has a positive return, since the total penalties over an episode of at most $H$ steps cannot surpass $R_{\text{success}}$; and 
        (ii) incentivize an efficient but balanced use of both modifications and queries. 
        During training, as previously mentioned, the query policy is masked to force periodic submits (\textit{PS}). 
        We scale the reward of these forced submits by $\lambda_{PS}\in[0,1]$, giving the training reward:
        \begin{equation} \label{eq:reward-forced}
        \resizebox{0.43\textwidth}{!}{$
            R(s_t,q_t,m_t) = \begin{cases}
                \lambda_{PS}\,R(s_t,q_t,m_t) & \text{if } t \equiv 0 \pmod{p}, \\
                R(s_t,q_t,m_t) & \text{otherwise.}
            \end{cases}
        $}
        \end{equation}
        We set $\lambda_{PS}=0$ for two reasons. First, early in training the agent is usually detected when it queries and can learn to avoid querying altogether; making these periodic queries zero-cost encourages it to explore querying. Second, once the agent is more capable, $\lambda_{PS}=0$ encourages the agent to voluntarily submit early to maximize reward.

    \subsection{\replicant Agent} \label{sec:replicant-agent}
        The general architecture and training pipeline that is used to make the \replicant agent is as follows, with the specific hyperparameters used in our experiments reported in Appendix~\ref{app:hyper-rep}. 
        A shared MLP encoder feeds three heads, each adding one further layer before its output: a 
        \textit{query} head over $\{\textsc{Submit},\textsc{Modify}\}$, a masked \textit{modify} head over the capability set $C$, and a \textit{critic} head estimating state value $V(s_t)$. 
        Training uses PPO~\cite{schulman2017proximal}, with the query policy loss applied at every step; 
        the modify policy loss applied \textit{only} on \textsc{Modify} steps, relying on the evasion reward being propagated through advantage estimation; and the critic is shared across both. 
        The query and modify heads share a common entropy coefficient, encouraging exploration over both \textit{when} to query and \textit{what} to add.
        Each update collects a batch of parallel episodes using complete-episode rollouts rather than a fixed number of steps, as the positive reward only arrives on episode termination.
        Because training yields a \textit{policy} $\pi_\theta$ rather than a fixed \ac{AE}, what \replicant learns is not bound to a single input or target: the same $\pi_\theta$ can be used to evade malware it has never seen and, as Section~\ref{sec:eval-transfer} shows, transfers to classifiers and feature spaces it was never trained against.

        \noindent\textbf{\replicantWB.}
        We also produce a white-box variant of \replicant (\replicantWB) for use in \ac{AT} and the robustness evaluations of the hardened targets in Sections~\ref{sec:eval-at}-\ref{sec:eval-temporal}. 
        Architecturally, \replicantWB drops the query head and submits after every modification, since query efficiency is not required in the white-box setting; it also uses the confidence score of the target, $\mathrm{clf}_{score}$, for rewards. 
        More specifically, the white-box rewards are as follows:
        \begin{equation} \label{eq:WB-reward-values}
        R_{\text{WB}}(s_t, c_{m_t})=\lambda_{WB}\,\big(\mathrm{clf}_{\text{score}}(s_t) - \mathrm{clf}_{\text{score}}(s_{t+1})\big)
        \end{equation}
        which rewards the reduction in the malware confidence of the target classifier, scaled by the hyperparameter $\lambda_{WB}$. 
        Since confidence deltas can be small, we scale them by $\lambda_{WB}=5$ for all targets. Empirically, this gives more signal to per-step rewards in training, allowing for faster convergence during training~\cite{bates2023reward}.
        Beyond architecture and reward, the rest of the \replicant formulation remains unchanged.

        \noindent\textbf{\atreplicant.}
        Our pipeline for \acf{AT} with \replicant (\atreplicant), used in the Sections~\ref{sec:eval-at}-\ref{sec:eval-temporal}, is as follows. 
        We extend the 15-round \ac{AT} methodology of~\cite{tsingenopoulos2024train}, where each round of \atreplicant trains a fresh \replicantWB policy against the classifier being hardened and collects every \ac{AE} generated over the course of that training run; the unique AEs collected are added to the training set of the classifier for retraining before the next round.
        As the classifier becomes harder to evade with each round~\cite{tsingenopoulos2024train}, a fixed training budget yields progressively fewer AEs, so we increase the number of policy updates as \ac{AT} progresses. 
        Empirically the number of new AEs collapses after five rounds under a fixed budget, whereas doubling the policy updates of \replicantWB every five rounds maintains AE production across all 15 rounds (details shown in Appendix~Table~\ref{tab:at-scaling}). 
        We therefore adopt this schedule as our default in all \atreplicant experiments.
        Simultaneously collecting AEs throughout training exposes the classifier to many distinct variants of every malware sample, unlike collection from an already-converged policy or via greedy composition.
        This diversity forces the classifier to adapt to a broader region of problem-space perturbations rather than overfitting to a single evasive strategy.

        \noindent\textbf{\ac{RAL}.}
        Finally, for performance and robustness under temporal drift (Section~\ref{sec:eval-temporal}), we propose \ac{RAL}, which extends \ac{AL} by running one \atreplicant round on the malware collected each \ac{AL} period and adds the resulting AEs to the \ac{AL} update set.

\section{Experimental Setup} \label{sec:setup}

    \noindent\textbf{Dataset.}
    In our experiments we use the \textit{Hypercube}~\cite{chow2025breaking} Android malware dataset, spanning $224{,}965$ samples from $2021$--$2024$, constructed to follow the current best practices for unbiased dataset creation: sampling from a single market (Google Play) using publication timestamps, at a statistically representative size, with a realistic malware distribution of $10\%$ and a VirusTotal Threshold of two (VTT2). 
    Additionally, we investigate the impact of VTT by also evaluating evasion on a variant of Hypercube sampled at a VTT of four (Hypercube$_{VTT4}$).
    We repeat every experiment: (i) over ten random seeds to account for variance~\cite{mcfadden2026sok}; and (ii) over two temporally distinct training periods ($2021$ \& $2022$) to avoid bias from temporal luck~\cite{chow2025breaking}. 

    \noindent\textbf{Training Sets.}
    For each seed, we perform a random stratified split to produce two equal sized training sets per training year: split~$a$ trains the surrogate and \replicant policies, and split~$b$ trains the target (e.g., $2021a$-vs-$2021b$ and $2022a$-vs-$2022b$). 
    To investigate transferability across temporal distributions for the black-box experiments (Sections~\ref{sec:eval-query}--\ref{sec:eval-transfer}), we also consider surrogate and target pairs across years (e.g., $2021a$-vs-$2022b$ and $2022a$-vs-$2021b$).\footnote{In the $2022a$-vs-$2021b$ setting, the surrogate overlaps the $2022$ test year, but excluding these overlapping samples changes \ac{ASR} by $<0.2$ percentage points (Appendix~Table~\ref{tab:overlap}); thus, we report over $2022$ for consistency and simplicity of discussion.}
    
    \noindent\textbf{Testing Sets.}
    We evaluate attack performance using malware from the year after target training (e.g., $2023$ for a $2022b$ target); which realistically mimics evasion by malware authors as they attempt to evade deployed detectors over time.
    For black-box experiments we consider a one year test period to keep temporal drift minimal, as drift would otherwise inflate \ac{ASR} (Appendix Table~\ref{tab:idood}).
    For the white-box experiments, we extend the test period to two years (e.g., $2023$--$2024$ for a $2022b$ target), as these evaluations measure the robustness gained through \ac{AT}, which should be assessed over a full deployment period rather than a single year ahead.
    For each seed and feature space, the evasion test set is the malware that \textit{every} classifier correctly detects ensuring that differences in \ac{ASR} reflect the attack rather than which malware each classifier happens to detect.

    \noindent\textbf{Variance.} 
    Following~\cite{mcfadden2026sok}, we repeat every experiment over ten random seeds and report the resulting variance through $95\%$ confidence intervals ($95\%$ CI) across seeds. Each seed repeats the network initializations, the policy training, the surrogate and target training splits, and, in turn, the test set.

    \noindent\textbf{Feature Spaces.} 
    We train all classifiers on the Drebin~\cite{Drebin}, APIGraph~\cite{zhang2020enhancing}, and RAMDA~\cite{ramda} feature spaces. 
    For Drebin, we follow the work of~\cite{kan2024tesseract,chow2025breaking,mcfadden2026drmd,demontis2017yes} and use the top 10k features based on feature-importance via linear model coefficients. 
    For APIGraph, following the original work~\cite{zhang2020enhancing}, we use Drebin for the non-API features and their clustering pipeline to enrich the API features.
    For Ramda, we use the original $379$ features as proposed in~\cite{ramda}. 
    In total, the Drebin, APIGraph, and RAMDA feature spaces contain $10{,}000$, $11{,}912$, and $379$ features, respectively. 
    We chose these three spaces for their differing sizes and overlap (Section~\ref{sec:AMD}), which allow us to evaluate transferability across varying degrees of alignment between the surrogate and target feature spaces.

    \noindent\textbf{Classifiers.} 
    We evaluate against the following seven classifiers: Linear SVM~\cite{Drebin}, SecSVM~\cite{demontis2017yes}, Random Forest~\cite{MaMaDroid}, DeepDrebin~\cite{deepDrebin}, RAMDA~\cite{ramda}, HCC~\cite{chen2023continuous}, and DRMD~\cite{mcfadden2026drmd}.
    We chose these seven classifiers because they span a diverse set of design principles, from margin maximization and ensemble voting to reconstruction, contrastive, and reward-weighted losses. 
    Therefore, this diversity allows us to evaluate transferability and performance across a representative set of architectures and optimization objectives in Android malware detection, giving a more generalizable measurement of the attacks considered.

    \noindent\textbf{Capability Set.} 
    Our capability set comprises gadgets harvested by prior problem-space work~\cite{pierazzi2020intriguing, cortellazzi2025intriguing, bostani2024effectiveness, jigsaw}, together with their feature-space effects we computed.
    Each gadget affects only a small set of features, so for each feature space we use the subset that changes at least one of its features, resulting in $3{,}143$ for Drebin, $4{,}804$ for APIGraph, and $1{,}739$ for RAMDA. 
    For cross-feature-space attacks, the attacker selects from the capability set applicable in their own feature space, and the target sees the effect of applying those gadgets in its own feature space. 
    As motivated and discussed in Section~\ref{sec:threat-model}, all attacks are performed under the same threat model and capability set for each setting, to ensure reported differences reflect attack strategy rather than access to different capabilities.

    \noindent\textbf{Baselines.} 
    We compare against three state-of-the-art problem-space Android malware evasion attacks, which all optimize samples individually:
    (i) \textit{APG}~\cite{pierazzi2020intriguing,cortellazzi2025intriguing}, white-box greedy capability selection by score contribution; 
    (ii) AdvDroidZero (\textit{ADZ})~\cite{he2023efficient}, gray-box tree search over a perturbation set using model confidence; 
    and, (iii) EvadeDroid (\textit{EvD})~\cite{bostani2024evadedroid}, gray-box greedy random capability selection that only keeps modifications that reduce target confidence.
    Following~\cite{mcfadden2026sok}, we also evaluate \textit{Rnd}, which uses the \replicant framework but selects actions uniformly \emph{at random}, to isolate the performance gains of policy learning.
    Hyperparameters for the baselines and classifiers can be found in the Appendix~\ref{app:hyper-clf}-\ref{app:hyper-rl}.

    \noindent\textbf{RL Design Ablation.}
    To isolate the gains provided by the architecture of our agent and its reward function over prior RL-based work, we compare against five alternative state-of-the-art RL agent designs for malware evasion:
    (i) \textit{MEME}~\cite{rigaki2023power}, model-based PPO;
    (ii) \textit{MAB-Malware} (MAB)~\cite{song2022mab}, Thompson-sampling multi-armed bandit;
    (iii) \textit{MERLIN} (MRLN)~\cite{quertier2022merlin}, deep Q-learning (DQN);
    (iv) \textit{AIMED-RL} (AIM)~\cite{labaca2021aimed}, distributional double DQN with NoisyNet exploration and prioritized replay;
    and (v) \textit{PSP-Mal} (PSP)~\cite{zhan2023psp}, dueling double DQN with Shapley-guided prioritized experience replay.
    As each of these five RL-based attacks were originally proposed for Windows PE malware evasion, we adapted them to Android by modifying them to use our environment, threat model, and capability set. 
    
    \noindent\textbf{Metrics.} 
    We report \ac{ASR}, specifically the percentage of malware that evaded detection at a fixed per sample budget of twenty queries, as our primary metric.
    We also report the average number of queries per sample (AvgQ) as a secondary metric for attacks to measure query-efficiency.
    We use the $F_1$ as our primary clean performance metric, using the Area Under Time (AUT) to measure the $F_1$ over time ($AUT(F_1)$) following~\cite{Tesseract,kan2024tesseract}.
    Furthermore, since metrics from a single training period are subject to temporal luck~\cite{chow2025breaking}, we report \ac{ASR} and $AUT(F_1)$ ($A$-$AUT(F_1)$~\cite{chow2025breaking}) averaged over both training periods ($2021b$ and $2022b$) unless otherwise specified (e.g., cross period evaluations in Sections~\ref{sec:eval-query}-\ref{sec:eval-transfer}).

    \noindent\textbf{Statistical Testing \& Table Reporting.} 
    For statistical testing, we use the Wilcoxon signed-ranks test, following~\cite{demvsar2006statistical}. 
    For reporting statistical significance in tables, the best result and any other value of statistical equivalence (at $p<0.01$) are reported in \textit{bold}. 
    The statistically second best values are reported in \textit{italics}. 
    For \textit{bold} values, the superscript denotes the statistical significance over the \textit{italics} value ($^{**}p<0.01$, $^{***}p<0.001$). Each result is reported as the mean $\pm$ $95\%$ confidence interval to capture variance across seeds.

\section{Evaluation} \label{sec:evaluation}

    Our evaluation is organized across five research questions to analyze the performance of \replicant in different scenarios. Through these scenarios and comparisons with multiple baselines, our evaluation demonstrates \replicant as a state-of-the-art evasion strategy.
    \begin{enumerate}[label=\textbf{RQ\arabic*.},leftmargin=*]

        \item \textit{Black-Box Evasion.} 
        Is \replicant the most effective and query-efficient black-box evasion attack when compared with prior problem-space attacks?

        \item \textit{Policy Transfer.}
        How do architectural, representational, and temporal differences affect \textit{policy} transfer?

        \item \textit{Adversarial Training.} 
        Does \atreplicant produce classifiers that are more robust to unseen attackers?
        
        \item \textit{Capability Drift.} 
        Does the robustness of \ac{AT} generalize to problem-space capabilities held out of training?
        
        \item \textit{Temporal Drift.} 
        Can a classifier retain the robustness of \ac{AT} while adapting to temporal drift with AL?
    
    \end{enumerate}

\subsection{Black-Box Evasion}\label{sec:eval-query}
    This section establishes \replicant as a state-of-the-art, query-efficient black-box evasion attack.
    We first present its query efficiency in the \textit{matched setting}, where the surrogate and target share a classifier, feature space, and training period, differing only in their training data ($42$ combinations per attacker).
    Then we show it is the strongest attack across \textit{all} surrogate/target settings, spanning seven classifiers, three feature spaces, and two training periods for both the surrogate and target ($1{,}764$ combinations per attacker).
    Lastly, we confirm that these conclusions are robust to both VTT and the distribution from which the test set is drawn.

    \noindent\textbf{Matched Setting.} 
    In the \textit{matched} setting, the surrogate and target classifiers share an architecture and feature space but are trained on separate training sets (Section~\ref{sec:setup}); with no difference between architecture or feature space, this setting represents the best-case scenario for a black-box attacker.
    The matched section of Table~\ref{tab:baselines} reports the mean \ac{ASR} for each attacker and target classifier, averaged over feature spaces and seeds.
    \replicant achieves a mean \ac{ASR} of $96.6\%$ using an average of $3.6$ queries per sample. 
    Thus outperforming every baseline in this setting, with minimal gap between surrogate and target, which is more closely aligned with the threat model of prior attacks.
    The strongest baseline, APG, reaches only $77.7\%$ while the ADZ, EvD, and random baselines sit at $61.5\%$--$69.7\%$.
    APG is statistically tied with \replicant on SVM, SecSVM and DeepDrebin, where ranking capabilities by their contribution to the surrogate decision function most closely aligns with the target boundary.
    The gap is largest on the RL-based DRMD, where APG collapses to $25.2\%$ against the $90.2\%$ of \replicant, followed by RAMDA and RF ($35.4$ \& $27.6$ points), with every other baseline statistically beaten on \textit{every} target.
    By feature space, the mean \ac{ASR} of \replicant is $98.8\%$ for Drebin, $98.3\%$ for APIGraph, and $92.5\%$ for the smaller $379$-feature RAMDA space.

    \begin{table}[t]
        \centering
        \caption{\textbf{Evasion Results}, attacker comparison against Android evasion baselines for both the \textit{matched} and \textit{all} settings, reporting mean \ac{ASR} and AvgQ for both. }
        \vspace{3pt}
        \label{tab:baselines}
        \adjustbox{max width=\columnwidth}{%
        \begin{tabular}{lccccc}
          \toprule
          \textbf{Matched} & \multicolumn{1}{c}{\textbf{\replicant}} & \multicolumn{1}{c}{\textbf{APG}} & \multicolumn{1}{c}{\textbf{ADZ}} & \multicolumn{1}{c}{\textbf{EvD}} & \multicolumn{1}{c}{\textbf{Rnd}} \\
          \cmidrule(lr){1-6}
          SVM        & $\mathbf{99.1}^{***}_{\pm 1.5}$ & $\mathbf{99.1}^{***}_{\pm 1.5}$ & $73.5_{\pm 3.6}$ & $\textit{84.0}_{\pm 2.8}$ & $74.2_{\pm 3.7}$ \\
          SecSVM    & $\mathbf{98.2}^{***}_{\pm 1.0}$ & $\mathbf{98.7}^{***}_{\pm 0.9}$ & $53.6_{\pm 4.0}$ & $\textit{73.5}_{\pm 1.9}$ & $54.3_{\pm 4.3}$ \\
          RF         & $\mathbf{93.3}^{***}_{\pm 1.7}$ & $65.8_{\pm 3.2}$ & $\textit{72.1}_{\pm 2.1}$ & $69.9_{\pm 2.3}$ & $70.7_{\pm 2.4}$ \\
          DeepDrebin & $\mathbf{96.5}^{***}_{\pm 2.1}$ & $\mathbf{94.5}^{***}_{\pm 2.3}$ & $56.0_{\pm 2.0}$ & $\textit{69.1}_{\pm 2.6}$ & $57.9_{\pm 2.1}$ \\
          RAMDA      & $\mathbf{99.9}^{***}_{\pm 0.1}$ & $64.5_{\pm 3.6}$ & $81.3_{\pm 3.8}$ & $72.0_{\pm 3.3}$ & $\textit{82.6}_{\pm 2.8}$ \\
          HCC        & $\mathbf{98.7}^{***}_{\pm 0.9}$ & $\textit{96.1}_{\pm 2.0}$ & $79.2_{\pm 2.5}$ & $85.2_{\pm 2.0}$ & $81.6_{\pm 3.0}$ \\
          DRMD       & $\mathbf{90.2}^{***}_{\pm 2.7}$ & $25.2_{\pm 2.3}$ & $14.5_{\pm 1.6}$ & $\textit{34.0}_{\pm 2.4}$ & $16.8_{\pm 1.9}$ \\
          \cmidrule(lr){1-6}
          \ac{ASR}   & $\mathbf{96.6}^{***}_{\pm 0.7}$ & $\textit{77.7}_{\pm 0.8}$ & $61.5_{\pm 1.1}$ & $69.7_{\pm 1.1}$ & $62.6_{\pm 2.0}$ \\
          AvgQ       & $\phantom{0}\mathbf{3.6}^{***}_{\pm 0.3}$ & $\phantom{0}\textit{6.5}_{\pm 0.2}$ & $10.6_{\pm 0.2}$ & $\phantom{0}9.9_{\pm 0.2}$ & $10.0_{\pm 0.2}$ \\
          \midrule
          \textbf{All Settings} & \multicolumn{1}{c}{\textbf{\replicant}} & \multicolumn{1}{c}{\textbf{APG}} & \multicolumn{1}{c}{\textbf{ADZ}} & \multicolumn{1}{c}{\textbf{EvD}} & \multicolumn{1}{c}{\textbf{Rnd}} \\
          \cmidrule(lr){1-6}
          \ac{ASR}   & $\mathbf{78.8}^{***}_{\pm 1.2}$ & $\textit{65.2}_{\pm 0.9}$ & $61.5_{\pm 1.2}$ & $56.6_{\pm 0.9}$ & $59.4_{\pm 1.1}$ \\
          AvgQ       & $\phantom{0}\mathbf{7.4}^{***}_{\pm 0.2}$ & $\phantom{0}\textit{9.3}_{\pm 0.2}$ & $10.6_{\pm 0.2}$ & $11.7_{\pm 0.2}$ & $10.4_{\pm 0.2}$ \\
          \bottomrule
        \end{tabular}}
      \end{table}

\begin{figure}[t]
        \centering
        \includegraphics[width=\linewidth]{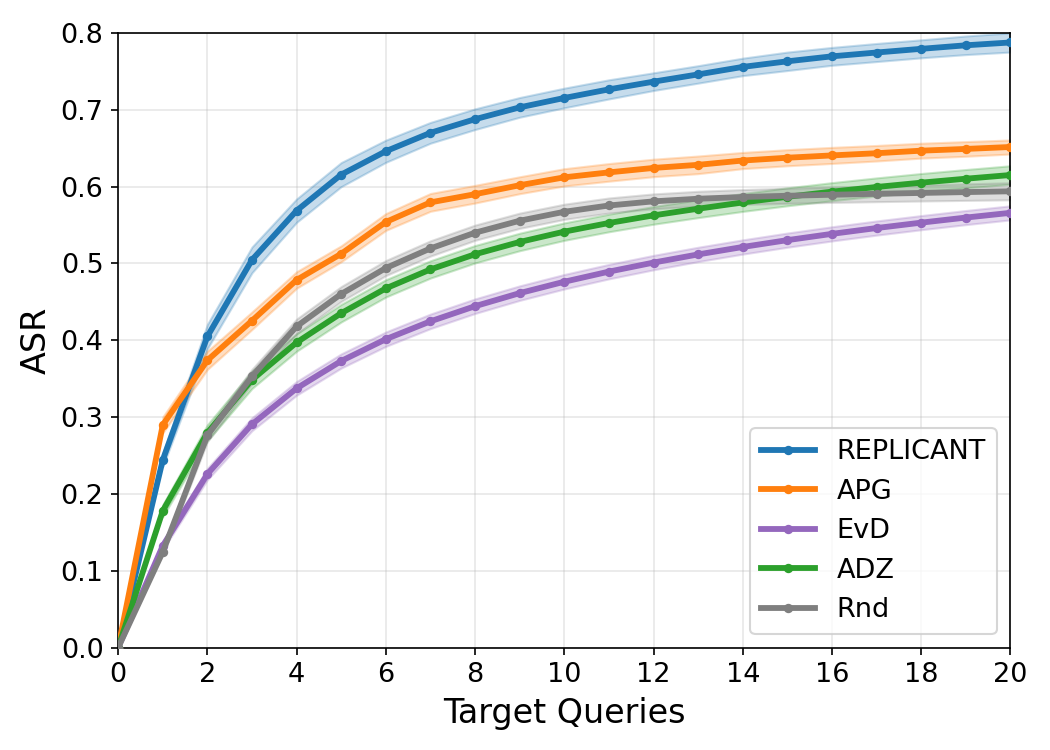}
        \caption{\textbf{Evasion Curves}, attack efficiency across all $1{,}764$ surrogate/target combinations and ten seeds. \replicant outperforms every baseline with the highest evasion rate per query.}
        \label{fig:allattack}
    \end{figure}

    \noindent\textbf{All Settings.} 
    In practice, an attacker rarely has a surrogate that matches the target, so we now evaluate \replicant across \textit{all} $1{,}764$ surrogate/target combinations, in which the surrogate and target may differ in classifier, feature space, training period, or all axes. 
    The ``all settings'' section of Table~\ref{tab:baselines} reports the mean \ac{ASR} and AvgQ across these settings, and Figure~\ref{fig:allattack} plots the full \ac{ASR} per query curve.
    \replicant achieves the highest evasion rate ($78.8\%$ \ac{ASR}), beating every Android baseline by $13.6$--$22.2$ \ac{ASR} points.
    One key advantage of \replicant is that it is truly black-box by design, learning only from binary decisions, whereas attacks such as ADZ and EvD rely on access to target confidence scores, a strong and less realistic assumption.
    Since the baselines use changes in confidence to steer capability selection, as the surrogate diverges from the target, its confidence degrades as a proxy for the decision boundary of the target, a similar failure mode to that of sample transfer evasion as found by~\cite{AttackTransfer}.    
    \replicant instead learns a \textit{policy} over the malware used to train the surrogate, thereby capturing the general task of evasion rather than the surrogate confidence topology.
    Therefore, we demonstrate that task-level knowledge transfers across settings more reliably than overfitting to confidence signals that the baselines rely on.

    \noindent\textbf{Query-Efficiency.}
    At a query budget of $20$, Table~\ref{tab:baselines} shows that \replicant is also the most query-efficient attacker by AvgQ, spending the fewest average queries per sample of any attack: $3.6$ in the matched setting and $7.4$ across all settings versus the $6.5$ and $9.3$ of APG, the Android strongest baseline.
    With Figure~\ref{fig:allattack} demonstrating that \replicant yields the highest \ac{ASR} per target query out of all attacks considered.

    \noindent\textbf{Policy Ablation.}
    Rnd uses the same MDP as \replicant, however, it selects \textit{when} to query and \textit{which} applicable capability to add at random. 
    Therefore, by evaluating Rnd, we can isolate the performance gain of the policy learned by \replicant from the base performance of the capability set and action masking.
    Figure~\ref{fig:allattack} illustrates that across all settings, Rnd matches the performance of both ADZ and EvD, while approaching the performance of APG at a budget of twenty queries.
    This demonstrates that: (i) the environment design of \replicant provides performance on par with prior attacks under our threat model; (ii) the performance gains over prior attacks are the result of our learned policy; and (iii) it exposes the weakness of relying on surrogate confidence scores for black-box attacks.

    \noindent\textbf{RL Design Ablation.}
    Although Android problem-space attacks are dominated by per-sample optimization, policy-based evasion strategies are more prevalent in the neighboring domain of Windows PE problem-space evasion.
    Therefore, to isolate the gains of our architecture and reward function, we compare against five alternative RL agent designs adapted to Android by using our environment, threat model, and capability set.  
    We find that in comparison to prior agent designs, \replicant improves the mean \ac{ASR} by $5.6$--$22.2$ points across \textit{all} $1{,}764$ settings, using $1.1$--$5.2$ less queries on average.
    Full results for both the \textit{matched} and \textit{all} settings can be found in Appendix~\ref{app:rl-ab}.

    \noindent\textbf{Datasets \& VTT.} 
    Our main results use the canonical Hypercube dataset, which samples the distribution and labels applications using VirusTotal threshold of 2 (VTT2). 
    To confirm our results are not an artifact of this threshold, we re-run the black-box experiments on a variant of Hypercube created for VTT4 (Hypercube$_{VTT4}$). 
    As a result of our threat model relying on surrogates for attack optimization and target agreement for test sets, the quality of data used to build both models is intrinsically linked with our evaluation. 
    Since VTT4 creates a higher standard for samples to be labeled as malware, it simplifies the detection task, dropping misclassification rates for $2022$ malware from $63.8\%$ to $46.9\%$.
    Thus VTT4 has a two-fold effect on our evaluation that makes it more favorable for the attacker: (i) it provides a larger test set as more malware are detected by all targets for a given seed and (ii) it produces more capable surrogates that provide a better proxy for the targets.
    These two factors result in a higher mean \ac{ASR} across all attacks; \replicant rises from $78.8\%$ (VTT2) to $94.1\%$ (VTT4), remaining the dominant attacker ($12.6\%$ better than the best baseline). 
    The full results for VTT4 are presented in Appendix Table~\ref{tab:baselines-vtt4}. 
    Therefore, since using VTT2 follows current best practice~\cite{chow2025breaking} and it is the more conservative setting for attack performance, our main results give a realistic measure of attack efficacy that is not inflated by higher labeling thresholds.

    \noindent\textbf{Evaluation Periods.} 
    The results discussed so far reflect the realistic \ac{ASR} against a deployed target, by evaluating the attacks on malware from the year following the target's training period (e.g., $2022$ malware for a $2021b$ target). 
    However, to confirm the attacks are also effective in-distribution, we additionally test on the malware from the surrogate split of each target's training year (e.g., $2021a$ malware for a $2021b$ target, and $2022a$ for a $2022b$ target). 
    The attack remains effective, with only a minor decline in performance; the mean \ac{ASR} of \replicant on in-distribution malware is $70.0\%$ compared to the $78.7\%$ of the main temporally withheld results. 
    Furthermore, when testing on malware from two and three years after the training periods, the mean \ac{ASR} rises to $89.4\%$ and $95.4\%$, respectively.
    The full breakdown of results for each testing period relative to the training period is reported in Appendix Table~\ref{tab:idood}.

    \begin{rqBox}
        \textbf{RQ1:} 
        \replicant is the most effective and query-efficient black-box attack across all $1{,}764$ settings, evading $78.8\%$ of malware within twenty queries.
        Thus beating every Android baseline by $13.6$--$22.2$ points (an improvement of $20.9\%$--$39.2\%$). 
    \end{rqBox}

\subsection{Transferring Policies}\label{sec:eval-transfer}
\begin{figure}[t]
      \centering
      \includegraphics[width=\linewidth]{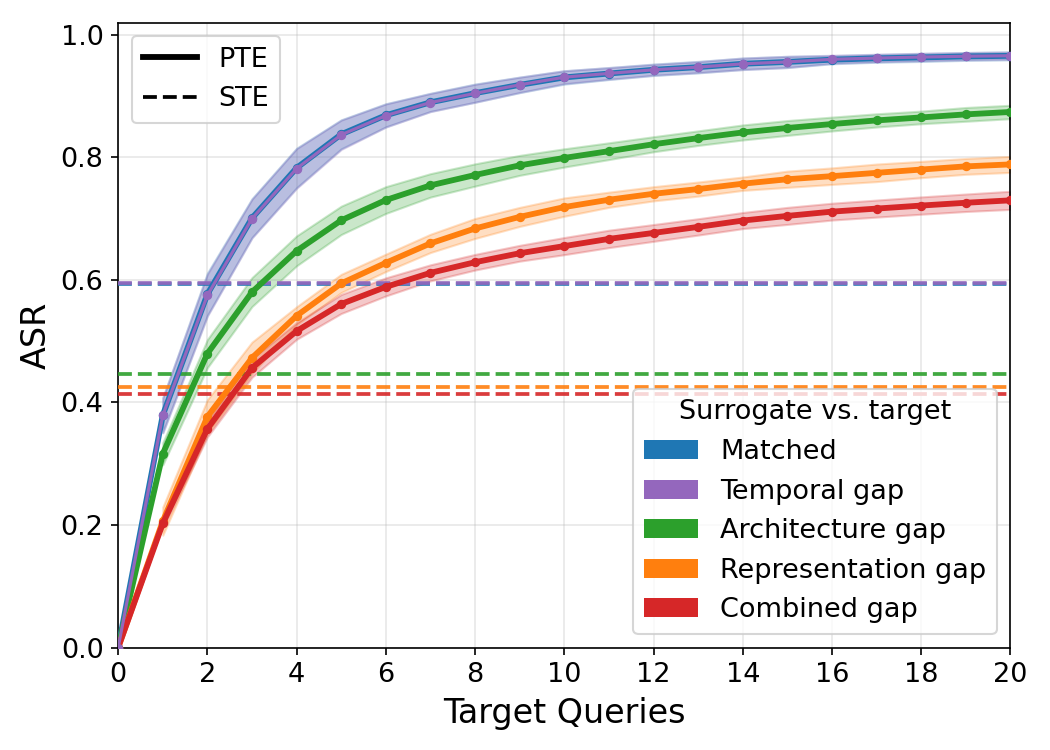}
      \caption{\textbf{Transferability Curves}, \acf{PTE} versus \acf{STE} for \replicant. Solid curves plot \ac{PTE} over queries and the dashed lines display the \ac{ASR} of \ac{STE}.}
      \label{fig:transfer}
    \end{figure}

    Having established \replicant as the strongest attacker in Section~\ref{sec:eval-query}, it is important to investigate how different surrogate-to-target gaps affect transferability. 
    Thus, we now provide a more detailed break down of the results from \textit{all settings} for \replicant in Table~\ref{tab:transfer}.
    Specifically, we consider four transferability axes: the temporal gap (when training periods differ), the architecture gap (when classifiers differ), the representation gap (when feature spaces differ), and the combined gap (when all differ).  
    Across these four axes, we analyze both \textit{\acf{PTE}}, where the policy learned by \replicant is deployed against the target, as it was done in Section~\ref{sec:eval-query}; and \textit{\acf{STE}}, where \replicant evades the surrogate, and those \acp{AE} are applied directly to the target.
    Therefore, both attacks use identical policies, capabilities, and surrogates; they only differ in whether the learned policy is employed as an active attacker or used to pre-compute \acp{AE}, as done in classical transfer attacks~\cite{AttackTransfer}.
    Table~\ref{tab:transfer} compares \ac{PTE} and \ac{STE} across the four axes we consider, and Figure~\ref{fig:transfer} plots their corresponding \ac{ASR} curves against target queries.

    \noindent\textbf{Temporal Gap.} 
    Matching the classifier and feature space but training the surrogate and target in \textit{different} periods (Figure~\ref{fig:transfer}, purple; per-period-pair breakdown in Appendix Figure~\ref{fig:transfer-temporal}), \ac{PTE} retains a mean \ac{ASR} of $96.6\%$, statistically indistinguishable from the matched setting. 
    Therefore, the learned policy is effectively invariant to the temporal gap; the only appreciable difference is between $2021b$ and $2022b$ targets as seen in Appendix Figure~\ref{fig:transfer-temporal}, which is a result of temporal luck~\cite{chow2025breaking} rather than the surrogate-target gap.

    \begin{table}[t]
      \centering
      \caption{\textbf{Transferability Results}, \acf{PTE} versus \acf{STE} for \replicant across the \textit{matched} setting and four axes of transferability within \textit{all} settings.}
      \vspace{3pt}
      \label{tab:transfer}
      \begin{tabular}{lccc}
        \toprule
        & \multicolumn{2}{c}{\textbf{\ac{PTE}}} & \\
        \cmidrule(lr){2-3}
        \textbf{Transfer Axes} & \multicolumn{1}{c}{\textbf{ASR}} & \multicolumn{1}{c}{\textbf{AvgQ}} & \multicolumn{1}{c}{\textbf{\ac{STE}}} \\
        \midrule
          Matched       & $\mathbf{96.6}^{***}_{\pm 0.7}$ & $3.6_{\pm 0.3}$ & $\textit{59.4}_{\pm 4.3}$ \\
          Temporal      & $\mathbf{96.6}^{***}_{\pm 0.6}$ & $3.6_{\pm 0.3}$ & $\textit{59.4}_{\pm 4.4}$ \\
          Architecture  & $\mathbf{87.4}^{***}_{\pm 1.1}$ & $5.8_{\pm 0.3}$ & $\textit{44.7}_{\pm 1.8}$ \\
          Representation& $\mathbf{78.8}^{***}_{\pm 1.3}$ & $7.6_{\pm 0.2}$ & $\textit{42.5}_{\pm 2.8}$ \\
          Combined      & $\mathbf{73.0}^{***}_{\pm 1.5}$ & $8.5_{\pm 0.2}$ & $\textit{41.3}_{\pm 1.9}$ \\
          \midrule
          All Settings  & $\mathbf{78.8}^{***}_{\pm 1.2}$ & $7.4_{\pm 0.2}$ & $\textit{43.3}_{\pm 1.8}$ \\
        \bottomrule
      \end{tabular}
    \end{table}

    \noindent\textbf{Architecture Gap.} 
    When the surrogate and target share the same feature space but differ in architecture (Figure~\ref{fig:transfer}, green; per-target breakdown in Appendix Figure~\ref{fig:transfer-arch}), transfer remains strong: \ac{PTE} evades $87.4\%$ of malware within twenty queries against $44.7\%$ for \ac{STE}, and overtakes \ac{STE} within roughly two queries.
    Given a shared representation, the learned effect that each modify action has on the feature space still holds for the target; the agent, therefore, retains an accurate model of how its capabilities change a sample. The differing architecture only relocates the boundary it must cross and the magnitude of the capability impact. 
    The architecture gap is thus paid foremost in queries: \ac{ASR} rises slowly against the targets whose boundaries differ most from the surrogate, most notably DRMD as shown in Appendix Figure~\ref{fig:transfer-arch}, the only RL-trained detector, whose categorically different training regime makes its boundary the least similar to the others.

    \noindent\textbf{Representation Gap.} 
    Keeping the architecture fixed while moving the surrogate and target into different feature spaces (Figure~\ref{fig:transfer}, orange; per-space-pair breakdown in Appendix Figure~\ref{fig:transfer-repr}) lowers \ac{ASR} to $78.8\%$, still far above the $42.5\%$ of sample transfer.
    The performance decay concentrates on pairs involving RAMDA especially when it is used as the surrogate as can be seen in Appendix Figure~\ref{fig:transfer-repr}. 
    A policy trained on RAMDA, the smallest feature space, and deployed against a richer target space has only learned capabilities that affect a small fraction of the target representation and must therefore rely either on that sub-region to influence the decision of the target or on incidental side-effects its capabilities produce elsewhere in the larger representation.
    Conversely, transferring an attack into a smaller representation, such as RAMDA, can render many of its modifications invisible to the target, as they are not captured by the representation of the target.
    This reduces the number of capabilities that have an tangible affect on the target, inducing a lower \ac{ASR} ceiling than the architecture gap, where the shared feature space leaves the full capability set intact. 
    Therefore, this reduction in \ac{ASR} reflects an underlying property of the representation gap rather than any intrinsic robustness of RAMDA, as shown by the high \ac{ASR} per query of RAMDA$\to$RAMDA in Appendix Figure~\ref{fig:transfer-repr}.    

    \noindent\textbf{Combined Gap.} 
    This is the hardest regime, where the surrogate differs from the target in architecture, feature space, and training period (Figure~\ref{fig:transfer}, red; per-target breakdown in Appendix Figure~\ref{fig:transfer-combined}).
    The costs of each transferability gap now compound: the policy pays in queries to push towards the shifted boundary and in expressiveness to compose capabilities in a representation it was not trained on, lowering the \ac{ASR} of \ac{PTE} to $73.0\%$, still far above the $41.4\%$ of \ac{STE}.
    Their combination also widens the spread across targets in Figure~\ref{fig:transfer-combined}: DRMD is the hardest target to evade, which is further compounded by the representation gap to produce the slowest per-query \ac{ASR} rise. Even so, \ac{PTE} reaches an \ac{ASR} of $41.1\%$ against it, whereas \ac{STE} reaches only $4.2\%$.

    \noindent\textbf{Policy over Sample Transfer.} 
    A transferred \textit{sample} is optimized to pass the decision boundary of the surrogate and relies on the alignment of decision boundaries.
    A transferred \textit{policy} instead encodes a \textit{distribution over actions} conditioned on the current representation of the sample.
    This provides \ac{PTE} with a distinct advantage that is exactly where \ac{STE} fails: as surrogate and target diverge, misaligned boundaries cause sample transfer to decay, while an adaptive policy re-queries until the decision boundary is crossed.
    Averaged over all settings, \ac{PTE} reaches $78.8\%$ \ac{ASR} against $43.3\%$ for \ac{STE}, and overtakes the \ac{ASR} of \ac{STE} within three queries on average.
    Therefore, when the surrogate and target agree on nothing but the distribution from which their training data was drawn, what transfers best is the learning of \textit{how to evade}, not specific evasive samples.

    \begin{rqBox}
        \textbf{RQ2:}
        As the gap between the surrogate and target grows, the \ac{ASR} of \replicant degrades gradually ($96.6\%{\rightarrow}73.0\%$). 
        Across all settings, \ac{PTE} reaches a mean \ac{ASR} of $78.8\%$ compared to the $43.3\%$ of \ac{STE} (an improvement of $82.0\%$).
        Demonstrating that policies consistently transfer across both architecture and representation, while \acp{AE} do not.
    \end{rqBox}

    \begin{table}[t]
      \centering
      \caption{\textbf{AT Results}, robustness comparison between \atreplicant and \apgat against both the \replicantWB and \apgWB attackers. Results report mean \ac{ASR}.}
      \vspace{3pt}
      \label{tab:at}
       \adjustbox{max width=\columnwidth}{%
      \begin{tabular}{lcccc}
        \toprule
        & \multicolumn{2}{c}{\textbf{\replicantWB}} & \multicolumn{2}{c}{\textbf{\apgWB}} \\
        \cmidrule(lr){2-3}\cmidrule(lr){4-5}
        \textbf{Classifier} & \multicolumn{1}{c}{\textbf{\atreplicant}} & \multicolumn{1}{c}{\textbf{\apgat}} & \multicolumn{1}{c}{\textbf{\atreplicant}} & \multicolumn{1}{c}{\textbf{\apgat}} \\
        \midrule
          SVM        & $\phantom{0}\mathbf{9.8}^{**}_{\pm 0.6}$ & $\textit{99.9}_{\pm 0.1}$ & $\mathbf{10.1}^{**}_{\pm 0.9}$ & $\textit{100.0}_{\pm 0.0}$ \\
          SecSVM    & $\phantom{0}\mathbf{8.9}^{**}_{\pm 2.8}$ & $\textit{48.4}_{\pm 4.2}$ & $\phantom{0}\mathbf{9.3}^{**}_{\pm 2.8}$ & $\textit{49.3}_{\pm 4.3}$ \\
          RF         & $\phantom{0}\mathbf{1.9}^{**}_{\pm 0.3}$ & $\textit{77.4}_{\pm 1.9}$ & $\phantom{0}\mathbf{3.7}^{**}_{\pm 0.3}$ & $\phantom{0}\textit{7.8}_{\pm 0.7}$ \\
          DeepDrebin & $\phantom{0}\mathbf{7.2}^{**}_{\pm 2.7}$ & $\textit{12.1}_{\pm 2.8}$ & $\mathbf{17.2}_{\pm 4.5}$ & $\mathbf{13.6}_{\pm 3.3}$ \\
          RAMDA      & $\mathbf{16.0}^{**}_{\pm 5.6}$ & $\textit{86.9}_{\pm 6.3}$ & $\mathbf{14.0}_{\pm 1.4}$ & $\mathbf{19.6}_{\pm 4.5}$ \\
          HCC        & $\mathbf{18.2}^{**}_{\pm 4.0}$ & $\textit{25.7}_{\pm 4.5}$ & $\mathbf{51.2}_{\pm 5.6}$ & $\mathbf{46.6}_{\pm 4.6}$ \\
          DRMD       & $\phantom{0}\mathbf{7.6}^{**}_{\pm 1.6}$ & $\textit{86.4}_{\pm 1.2}$ & $\phantom{0}\mathbf{8.8}^{**}_{\pm 1.8}$ & $\textit{12.2}_{\pm 1.8}$ \\
          \midrule
          Mean       & $\phantom{0}\mathbf{9.9}^{***}_{\pm 1.3}$ & $\textit{62.4}_{\pm 1.8}$ & $\mathbf{16.3}^{***}_{\pm 1.2}$ & $\textit{35.6}_{\pm 1.1}$ \\

        \bottomrule
      \end{tabular}
      }
    \end{table}

\subsection{Adversarial Training} \label{sec:eval-at}
    The robustness gained from adversarial training depends on the quality of the attacker~\cite{madry2017towards}, so we harden with the strongest attacker from each attack type (i.e., policy-based and per-sample optimization). 
    Therefore, motivated by the results of Section~\ref{sec:eval-query}, we use \replicant for learned policies and APG for per-sample optimization to provide the most informative comparison.
    Since \ac{AT} relies on successful evasions and classifier robustness is best evaluated under the strongest threat model~\cite{carlini2019evaluating}, we conduct all experiments in this section using \textit{white-box} attacks ($_{WB}$).
    As we are not considering transferability, and evasion rates are similar between the feature spaces in the matched setting (Section~\ref{sec:eval-query}), we focus this analysis on the Drebin feature space because it is the most widely used representation~\cite{Drebin}.
    \atreplicant and \apgat follow the pipeline presented in Section~\ref{sec:replicant-agent}, however, \apgat collects AEs during its per-sample rollouts as it does not have a training phase.

    \begin{figure}[t]
      \centering
      \includegraphics[width=\linewidth]{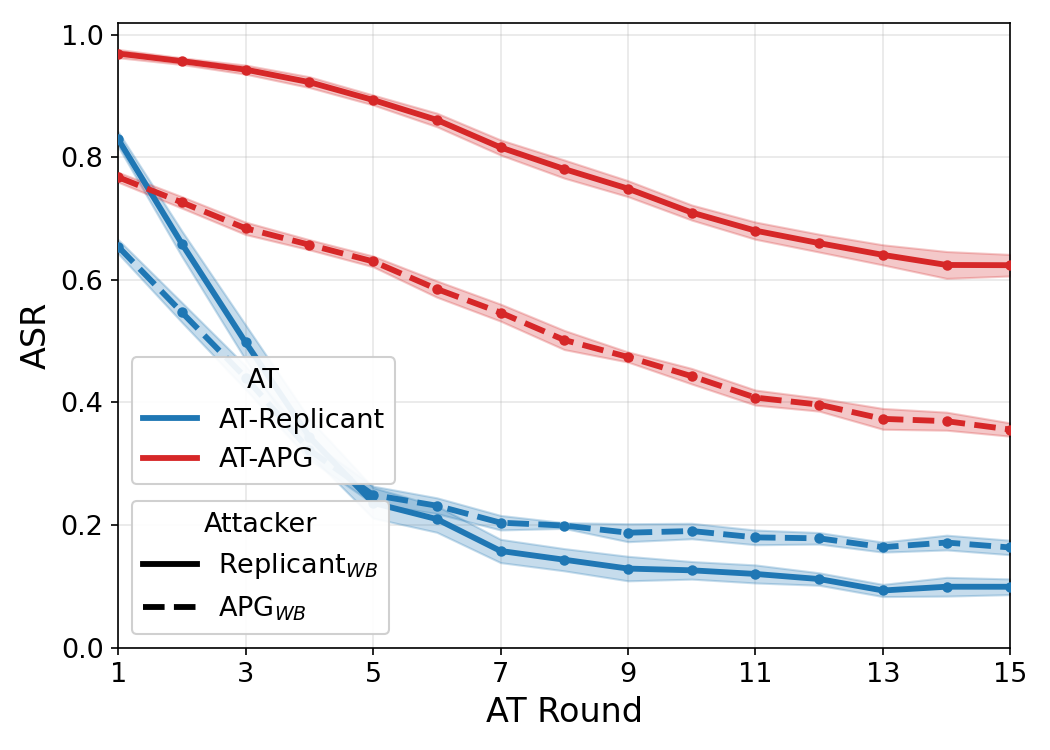}
      \caption{\textbf{AT Curves}, \ac{ASR} across both attackers and \ac{AT} pipelines for each of the 15 \ac{AT} rounds. The colors denote the \ac{AT} pipeline and line style denotes the attacker.}
      \label{fig:at}
    \end{figure}
    
    \noindent\textbf{Robustness Evaluation.} 
    We evaluate each classifier trained by both \ac{AT} pipelines against both \replicantWB and \apgWB attacks.
    Table~\ref{tab:at} reports the \ac{ASR} of the white-box attackers against both \ac{AT} pipelines after 15 rounds and Figure~\ref{fig:at} plots the \ac{ASR} across rounds.
    After 15 rounds, \atreplicant reduces the mean \ac{ASR} to $9.9\%$ and $16.3\%$ for \replicantWB and \apgWB, respectively. 
    Therefore, \atreplicant produces classifiers that are robust against both the attack it was trained on and an unseen attacker.
    Conversely, \apgat primarily reduces the mean \ac{ASR} of \apgWB ($35.6\%$) while remaining vulnerable to \replicantWB ($62.4\%$).
    As shown in Figure~\ref{fig:at}, this asymmetry holds throughout training: both attackers' \ac{ASR} decay similarly across subsequent rounds of \atreplicant, while the performance of \replicantWB reduces more slowly than \apgWB across successive rounds of \apgat.

    \noindent\textbf{\atreplicant Generalization.}
    APG is greedy and deterministic, thus each round naturally focuses on a small sub-set of capabilities and limits the diversity of \acp{AE} produced.
    As a result, \replicantWB learns capability sequences outside from the targeted paths \apgat trained on, and achieves a mean \ac{ASR} of $62.4\%$.
    Conversely, the robustness learned by \atreplicant generalizes better as it collects diverse AEs during \textit{stochastic policy training}, providing significantly more coverage of attack paths. 
    Therefore, the \textit{coverage} of possible paths modulates how broadly the robustness gained by \ac{AT} can generalize.

    \noindent\textbf{Clean Performance.} 
    We also report the clean performance (mean first month $F_1$ and A-AUT($F_1$)) of the classifiers produced by both \ac{AT} pipelines in Appendix Table~\ref{tab:at-clean}.
    \atreplicant has a more adverse effect than \apgat to the initial $F_1$ performance in the first test month (mean $F_1$ drop across classifiers of $-5.6$ versus $-3.1$).
    However, when considering performance-over-time, both \ac{AT} pipelines increase mean $A$-$AUT(F_1)$ in four of the seven total classifiers, with \atreplicant hardened classifiers improving mean A-AUT($F_1$) by $2.8$ compared to the $1.1$ gain of \apgat.
    Therefore, although \ac{AT} reduces initial performance, the new malware variants for training created by \atreplicant mildly improve the generalization of most classifiers as samples start to drift over time.

    \begin{table}[t]
      \centering
      \caption{\textbf{Clean Performance}, for the base, \atreplicant, and \apgat trained classifiers over the first evaluation month ($F_1$) and $A$-$AUT(F_1)$.}
      \vspace{3pt}
      \label{tab:at-clean}
        \adjustbox{max width=\columnwidth}{%
      \begin{tabular}{lcccccc}
        \toprule
        & \multicolumn{3}{c}{First-Month $\mathbf{F_1}$} & \multicolumn{3}{c}{3-Year $\mathbf{A}$\textbf{-}$\mathbf{AUT(F_1)}$} \\
        \cmidrule(lr){2-4}\cmidrule(lr){5-7}
        \multicolumn{1}{c}{\textbf{Classifier}} & \multicolumn{1}{c}{\textbf{Base}} & \multicolumn{1}{c}{\textbf{\atreplicant}} & \multicolumn{1}{c}{\textbf{\apgat}} & \multicolumn{1}{c}{\textbf{Base}} & \multicolumn{1}{c}{\textbf{\atreplicant}} & \multicolumn{1}{c}{\textbf{\apgat}} \\
        \midrule
          SVM        & $\mathbf{81.7}^{**}_{\pm 0.2}$ & $73.7_{\pm 0.5}$ & $\textit{80.7}_{\pm 0.5}$ & $\mathbf{25.8}_{\pm 0.7}$ & $\mathbf{25.0}_{\pm 0.5}$ & $\mathbf{25.7}_{\pm 0.6}$ \\
          SecSVM    & $\mathbf{78.7}^{**}_{\pm 0.6}$ & $60.8_{\pm 3.2}$ & $\textit{68.3}_{\pm 3.6}$ & $24.7_{\pm 0.6}$ & $\mathbf{32.2}^{**}_{\pm 0.8}$ & $\textit{28.9}_{\pm 0.7}$ \\
          RF         & $\mathbf{83.3}^{**}_{\pm 0.3}$ & $\textit{81.8}_{\pm 0.6}$ & $\mathbf{84.0}^{**}_{\pm 0.4}$ & $23.5_{\pm 0.4}$ & $\mathbf{31.8}^{**}_{\pm 0.3}$ & $\textit{25.0}_{\pm 0.5}$ \\
          DeepDrebin & $\textit{79.4}_{\pm 1.0}$ & $\mathbf{80.1}^{**}_{\pm 0.7}$ & $\mathbf{81.3}^{**}_{\pm 0.4}$ & $\textit{24.7}_{\pm 1.5}$ & $\mathbf{29.0}^{**}_{\pm 1.1}$ & $\mathbf{28.4}^{**}_{\pm 0.7}$ \\
          RAMDA      & $\mathbf{80.9}_{\pm 0.6}$ & $\mathbf{79.6}_{\pm 0.4}$ & $\mathbf{77.5}_{\pm 3.3}$ & $\textit{23.9}_{\pm 1.1}$ & $\mathbf{28.5}^{**}_{\pm 0.9}$ & $\mathbf{27.9}^{**}_{\pm 1.1}$ \\
          HCC        & $\mathbf{82.3}_{\pm 0.8}$ & $\mathbf{81.7}_{\pm 0.6}$ & $\mathbf{77.6}_{\pm 3.8}$ & $\mathbf{28.6}^{**}_{\pm 1.2}$ & $\textit{26.3}_{\pm 1.1}$ & $25.1_{\pm 1.4}$ \\
          DRMD       & $\mathbf{76.8}^{**}_{\pm 1.3}$ & $65.9_{\pm 1.8}$ & $\textit{71.3}_{\pm 1.4}$ & $\mathbf{36.0}^{**}_{\pm 0.5}$ & $\textit{33.8}_{\pm 0.8}$ & $33.4_{\pm 0.4}$ \\
          \midrule
          Mean       & $\mathbf{80.4}^{***}_{\pm 0.6}$ & $74.8_{\pm 1.9}$ & $\textit{77.3}_{\pm 1.5}$ & $26.7_{\pm 1.0}$ & $\mathbf{29.5}^{***}_{\pm 0.8}$ & $\textit{27.8}_{\pm 0.7}$ \\
        \bottomrule
      \end{tabular}
      }
    \end{table}

    \begin{rqBox}
        \textbf{RQ3:} 
        \atreplicant provides significantly more robustness to both attackers ($9.9\%$--$16.3\%$) compared to \apgat, which only reduces mean \ac{ASR} to $35.6\%$ and $62.4\%$ for \apgWB and \replicantWB, respectively.
        This demonstrates that the \textit{diversity} of AEs collected during policy training provides better coverage of the adversarial perturbation space (compared to sample-based optimizations), improving robustness and generalization to an unseen attacker.
    \end{rqBox}

\subsection{Capability Drift} \label{sec:eval-capdrift}
\begin{figure}[t]
      \centering
      \includegraphics[width=\linewidth]{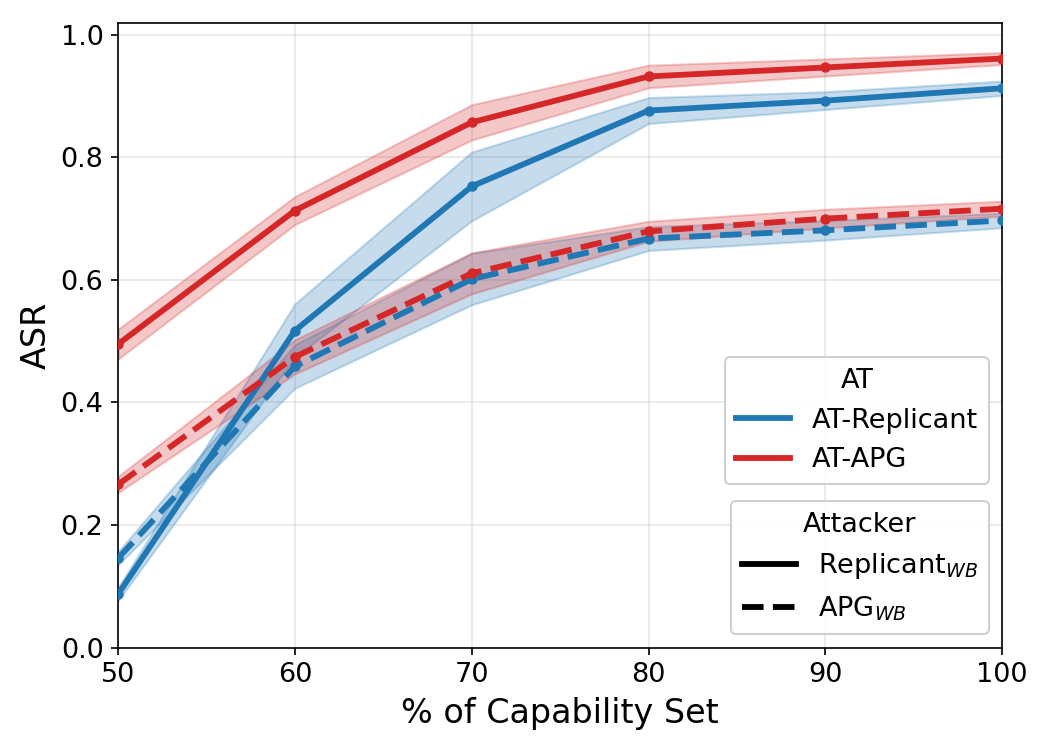}
      \caption{\textbf{Capability Drift Curves}, \ac{ASR} for the attackers as the capability set grows from reduced set ($50\%$ of original) to the full capability set. 
      }
      \label{fig:capdrift}
    \end{figure}

    In this section we test whether the robustness learned by the \ac{AT} pipelines generalizes to \textit{unseen problem-space capabilities}, which we refer to as capability drift. 
    Therefore, we split the capability set in two: (i) the capability set used by both the \ac{AT} and the initial attacker; and, (ii) the held-out attacker capabilities used to measure generalization to capability drift.
    We perform \ac{AT} and the subsequent robustness evaluation as was done in Section~\ref{sec:eval-at}; however, attackers are evaluated six separate times, starting with the initial capability set, and then progressively adding $10\%$ of the total capabilities back in each subsequent evaluation. 
    Figure~\ref{fig:capdrift} plots the \ac{ASR} as the capability set of the attacker drifts, with capabilities measured as a percentage of the original capability set.

    \noindent\textbf{Robustness Decay.}
    When reducing the \ac{AT} capability set (attacker and defender have the same capabilities), results reflect the findings of Section~\ref{sec:eval-at}: \atreplicant shows robustness against both \replicantWB ($8.7\%$) and \apgWB ($14.6\%$), whereas \apgat primarily against \apgWB ($26.6\%$ versus the $49.5\%$ of \replicantWB).
    However, as capability drift increases so does the mean \ac{ASR} of \replicantWB, with the full capability set reaching $91.3\%$ (\atreplicant) and $96.1\%$ (\apgat).
    Therefore, as the attacker set changes, the two \ac{AT} pipelines demonstrate similar robustness to \apgWB; however, \atreplicant remains \textit{consistently} more robust than \apgat to the \replicantWB attacker at \textit{every} stage of capability drift (mean \ac{ASR} reduction of $-40.8$ to $-4.9$ percentage points).
    This demonstrates that although \atreplicant generalizes better to capability drift than \apgat, \ac{AT} has natural limitations with respect to the capability set it was trained against.
    Therefore, in practical deployments just as classifiers need \ac{AL} to adapt to temporal drift, \ac{AT} must also adapt to capability drift.

    \begin{rqBox}
        \textbf{RQ4:}
        \atreplicant is more robust than \apgat at every stage of capability drift; however, the robustness of both pipelines decays as capabilities increase.
        Demonstrating that although \atreplicant generalizes better across capability drift, \ac{AT} remains bound to its training capabilities and requires continuous adaptation to stay ahead of evolving threats.
    \end{rqBox}

\subsection{Temporal Drift} \label{sec:eval-temporal}
    Over time both goodware and malware evolve, without adaptation, the performance of a trained detector decays as new samples drift further from the training distribution.
    Therefore, in this section, we investigate the impact that temporal drift and its mitigation, \acf{AL}~\cite{settles2009active}, have on both performance and robustness over time.
    We compare five different pipelines for the classifiers against fixed \replicant attackers (under both the white-box and black-box settings): 
    (i) \textit{Base}, the static classifier from its respective training period; 
    (ii) \textit{AL}, which performs monthly \ac{AL} updates with $400$ samples selected from that period using uncertainty-based selection, following~\cite{Tesseract,kan2024tesseract,mcfadden2026drmd}; 
    (iii) \textit{\ac{AT}}, the \atreplicant pipeline from Section~\ref{sec:eval-at}; 
    (iv) \textit{\ac{AT}\&\ac{AL}}, the \atreplicant hardened classifiers with \ac{AL} across the test periods; 
    and, (v) \textit{\acf{RAL}}, which we proposed in Section~\ref{sec:replicant-agent}.

    \begin{table}[t]
      \centering
      \caption{\textbf{Temporal Drift Results}, clean performance and robustness over time reported as A-AUT($F_1$) and A-AUT(ASR), respectively. The WB (\replicantWB) and BB (\replicant) attacks are fixed policies from their training period.}
      \vspace{3pt}
      \label{tab:temporal}
      \begin{tabular}{lccc}
        \toprule
        & & \multicolumn{2}{c}{\textbf{A-AUT(ASR)}} \\
        \cmidrule(lr){3-4}
        \textbf{Setting} & \multicolumn{1}{c}{\textbf{A-AUT(}$\mathbf{F_1}$\textbf{)}} & \multicolumn{1}{c}{\textbf{WB}} & \multicolumn{1}{c}{\textbf{BB}} \\
        \midrule
           Base    & $26.7_{\pm 0.5}$ & $99.9_{\pm 0.0}$ & $93.9_{\pm 0.9}$ \\
           \ac{AL}      & $\mathbf{49.8}^{***}_{\pm 0.1}$ & $98.2_{\pm 0.2}$ & $88.5_{\pm 0.7}$ \\
           \ac{AT}      & $29.5_{\pm 0.4}$ & $\phantom{0}\mathbf{0.1}^{***}_{\pm 0.1}$ & $\phantom{0}\mathbf{0.4}^{***}_{\pm 0.1}$ \\
           \ac{AT}\&\ac{AL}  & $\textit{47.9}_{\pm 0.2}$ & $19.4_{\pm 2.1}$ & $\textit{20.6}_{\pm 1.2}$ \\
           \ac{RAL}     & $\mathbf{49.0}^{***}_{\pm 0.5}$ & $\phantom{0}\textit{0.6}_{\pm 0.3}$ & $\phantom{0}\mathbf{0.8}^{***}_{\pm 0.4}$ \\
        \bottomrule
      \end{tabular}
    \end{table}

    \noindent\textbf{Performance \& Robustness over Time.}
    Table~\ref{tab:temporal} demonstrates that drift adaptation (${\uparrow}$A-AUT($F_1$)) and adversarial robustness (${\downarrow}$A-AUT(ASR) WB \& BB) are largely orthogonal, with no prior strategy improving \textit{both} metrics over the three-year evaluation.
    The \textit{Base} classifiers perform poorly on both, losing clean performance to drift ($26.7$) while the fixed attacker policies evade almost completely ($99.9\%$ WB, $93.9\%$ BB).
    \textit{\ac{AL}} recovers the most clean performance of any configuration ($49.8$) but, as it never exposes the classifiers to AEs in training, the adaptation only has a minor reduction on attack performance ($98.2\%$ WB, $88.5\%$ BB); \textit{\ac{AT}} conversely suppresses the attackers to near-zero \ac{ASR}, however, provides only minor robustness to temporal drift ($29.5$).
    Combining \textit{\ac{AT}\&\ac{AL}} improves performance with regards to temporal drift at the cost of adversarial robustness ($19.4\%$ WB, $20.6\%$ BB).
    Finally, only our \textit{\ac{RAL}}, with its combination of clean and \acp{AE} for retraining, improves on both axes, matching the clean performance of \ac{AL} ($49.0$ versus $49.8$) while preserving the adversarial robustness from \ac{AT} ($0.6\%$ WB versus $0.1\%$ WB).
    Therefore, just as \ac{AL} is needed to adapt to temporal drift, \ac{AT} must continually adapt alongside it, as neither \ac{AL} nor \ac{AT} alone is sufficient.

    \begin{rqBox}
        \textbf{RQ5:}
        Directly applying \ac{AL} on hardened classifiers improves clean performance at the cost of robustness, however, \ac{RAL} provides performance-over-time while maintaining robustness; matching the performance of \ac{AL} ($49.0$) and the robustness of \ac{AT} ($0.6\%$ WB).
    \end{rqBox}

\section{Limitations}\label{sec:limitations}
    \replicant addresses the \textit{strategy} of problem-space evasion given a fixed capability set, rather than the design of the capability set itself. 
    This scope is deliberate, as conflating newly proposed capability sets and strategies hinders direct comparability and gain attribution~\cite{mcfadden2026sok}. 
    Therefore, in each scenario, we use the same capability set and threat model across all attackers. 
    As a result, the \acp{ASR} we report are relative to the capability set and dataset used rather than an absolute upper bound on problem-space evasion, and determining the strongest capability set remains an important open problem, with works such as~\cite{scano2026droidbreaker} exploring that line of research. 
    However, \replicant is not intrinsically tied to the capabilities used in this work and can be applied to new capability sets as they are developed.  
    For datasets, we only use the Android malware dataset of Hypercube~\cite{chow2025breaking}; we made this choice deliberately to avoid the experimental bias of other datasets, as~\cite{chow2025breaking} constitutes the current standard for unbiased dataset construction.
    Although there are some natural limitations in our study, our work still constitutes the largest and most comprehensive study of malware evasion to date, comprising $379{,}680$ attack evaluations (Appendix~\ref{sec:app-scale}) in order to isolate improvements in attacker strategy.

\section{Related Work}

    \noindent\textbf{Malware Evasion.} 
    In \ac{AML}, \textit{feature-space} attacks perturb the input vector of the classifier directly. They were initially formalized for security by~\cite{biggio2013evasion} and then popularized in computer vision by~\cite{goodfellow2014explaining} and~\cite{carlini2017towards}. 
    In the domain of (Android) malware detection~\cite{deepDrebin} uses saliency maps to perturb features, causing the inverse mapping problem: perturbations do not necessarily correspond to a real program~\cite{pierazzi2020intriguing}. 
    \textit{Problem-space} attacks avoid this problem by modifying the application itself: \cite{pierazzi2020intriguing} introduces the realizability, plausibility, preprocessing-robustness, and semantic-preservation constraints, in addition to APG. 
    Similarly, \cite{bostani2024evadedroid} and \cite{he2023efficient} harvest and transplant code gadgets under gray-box access. 
    While many works develop their own capability sets, such as~\cite{scano2026droidbreaker}, \replicant presents a novel way to \textit{learn} evasion strategies. 
    Conversely, \ac{DRL}-based problem-space evasion for malware is instead a relatively nascent field in the Android domain and is primarily concentrated in the neighboring Windows PE domain including the approaches we adapted~\cite{zhan2023psp,rigaki2023power,song2022mab,quertier2022merlin,labaca2021aimed} and other work such as \cite{tsingenopoulos2022adaptive}, which evades dynamic detectors. 

    \noindent\textbf{Black-Box Attacks.} 
    Boundary~\cite{brendel2017decision} and HopSkipJumpAttack~\cite{chen2020hopskipjumpattack} popularized decision-based attacks in vision, and \cite{debenedetti2024evading} extended querying attacks to minimize detected queries in the process, while \cite{dong2021query} and \cite{cheng2024efficient} improve query efficiency by using transfer-based and Bayesian priors. 
    \cite{rosenberg2020query, digregorio2024tarallo} target sequence-based classifiers and \cite{rashid2023malprotect} propose a stateful defense against this type of attack. 
    Finally, \cite{tsingenopoulos2025adaptive} use DRL to optimize both the offensive and defensive sides during query-based attacks. Contrary to decision/query-based, transferability attacks optimize against a surrogate and rely on decision-boundary alignment; \cite{AttackTransfer} provide the foundational analysis of why and when adversarial examples transfer, attributing success to gradient and boundary alignment between surrogate and target. 
    \replicant instead combines transfer and query based attacks into a single threat model. 
    In doing so it learns a policy that transfers across classifiers and feature spaces while minimizing queries used. 

    \noindent\textbf{Adversarial Training.} 
    Designing malware detectors that are naturally harder to evade dates back to \cite{demontis2017yes} and later \cite{ramda}, which manipulates features and their input weight for robustness. 
    Yet such classifiers can still be evaded, adversarial training was thus introduced to retrain classifiers with malicious samples to reduce evasion rates~\cite{madry2017towards}. 
    More recent work moves adversarial training into the problem space: \cite{zhang2025fighting} and \cite{tsingenopoulos2024train} introduce a continuous attack-and-retrain pipeline for Android and Windows PE detectors, respectively. 
    \cite{lucas2023adversarial} study adversarial training for raw-bytes classifiers, investigating the impact of problem-space transformations as well as AEs that are not problem-space compliant. 
    Although robustness and temporal drift are typically studied in isolation, we draw on the temporal-drift and active-learning literature motivating our \textit{RQ5}. 
    Tesseract~\cite{Tesseract, kan2024tesseract} and \cite{DosDonts} establish the evaluation methodology we adopt, \cite{mcfadden2024impact, mcfadden2023poster} study the interaction between \ac{AL} and adversarial data poisoning. 
    Our work complements this literature by showing that neither adversarial training nor \ac{AL} alone is sufficient, and that only their interleaving (\ac{RAL}) preserves both properties simultaneously.

\section{Conclusion}
    We present \replicant, a hierarchical \ac{DRL} framework that reframes problem-space malware evasion as a \textit{policy} learning problem rather than a per-sample optimization problem.
    Given a set of problem-space capabilities and a surrogate, \replicant learns both \textit{what} capability to apply and \textit{when} to query, yielding a query-efficient black-box attack that transfers across classifier architectures and feature spaces.
    Across seven classifiers, three feature spaces, and two training periods, \textit{transferring the \replicant policy} evades, on average over all settings, $78.8\%$ of samples within twenty queries. 
    Conversely, \textit{sample transfer} succeeds in $43.3\%$ of the cases, while prior Android attacks achieve $65.2\%$--$56.6\%$.
    For adversarial training, \replicantWB drives residual attack success below $17\%$ against both itself and APG, while staying more robust than \apgat to unseen capabilities.
    Furthermore, we show that maintaining robustness while maximizing clean performance across three years of drifting samples requires interleaving both \ac{AL} and \ac{AT}.
    Together, these results establish learned policies as a state-of-the-art foundation for both attacking and adversarially training malware detectors.

\bibliographystyle{plain}
\bibliography{references}

\appendix

\section{Common Acronyms}

\begin{acronym}
    \acro{DRL}{Deep Reinforcement Learning}
    \acro{MDP}[MDP]{Markov Decision Process}
    \acro{ML}{Machine Learning}
    \acro{AML}{Adversarial Machine Learning}
    \acro{AE}[AE]{Adversarial Sample}
    \acro{AT}{Adversarial Training}
    \acro{AL}{Active Learning}
    \acro{RAL}{Robust \ac{AL}}
    \acro{PTE}{Policy Transfer Evasion}
    \acro{STE}{Sample Transfer Evasion}
    \acro{ASR}{Attack Success Rate}
\end{acronym}

\section{Proximal Policy Optimization} \label{app:ppo}
This appendix expands the three components of the PPO objective stated in Equation~\ref{eq:ppo-loss}. PPO~\cite{schulman2017proximal} is an actor-critic, on-policy algorithm that operates as follows: at each iteration, $T$ environment steps are collected under the current policy $\pi_{\theta_{\mathrm{old}}}$ into a rollout buffer $\mathcal{D} = \{(s_t, a_t, r_t, d_t)\}_{t=0}^{T-1}$ (with $d_t \in \{0,1\}$ denoting if the episode has terminated), and $\theta$ is then updated for $K$ epochs of minibatch stochastic gradient descent on $\mathcal{L}^{\mathrm{PPO}}(\theta)$ before a fresh rollout is collected. PPO constrains updates with probability-ratio clipping and has become the de facto standard in the literature. 

\subsection{Clipped Surrogate Objective} \label{app:ppo-policy}
    To enable multiple epochs over a single rollout, PPO uses the surrogate
    \begin{equation}
        \mathbb{E}_t\!\big[\rho_t(\theta)\,\hat{A}_t\big],
        \qquad
        \rho_t(\theta) \;=\; \frac{\pi_\theta(a_t \!\mid\! s_t)}{\pi_{\theta_{\mathrm{old}}}(a_t \!\mid\! s_t)},
    \end{equation}
    where $\pi_{\theta_{\mathrm{old}}}$ is used during rollout and $\rho_t(\theta)$ is the per-step probability ratio between the policies. Maximizing the surrogate without constraining $\pi_\theta$ causes too much drift with respect to $\pi_{\theta_{\mathrm{old}}}$ and degrades performance. Therefore, PPO controls this drift via a ratio-clipping mechanism:
    \begin{equation}\label{eq:ppo-clip}
        \mathcal{L}^{\mathrm{P}}(\theta) \;=\; \mathbb{E}_t\!\left[
        \min\!\Big(\rho_t(\theta)\,\hat{A}_t,\;
        \mathrm{clip}\!\big(\rho_t(\theta),\, 1-\epsilon,\, 1+\epsilon\big)\,\hat{A}_t\Big)
        \right],
    \end{equation}
     where $\epsilon \in (0,1)$ is the clip range. The clip bounds $\rho_t$ to $[1-\epsilon,\,1+\epsilon]$, and the min operator removes the incentive to move the ratio further. When $\hat{A}_t > 0$, the contribution is capped at $(1+\epsilon)\hat{A}_t$, so raising $\rho_t$ beyond $1+\epsilon$ yields no further gain; when $\hat{A}_t < 0$, it is floored at $(1-\epsilon)\hat{A}_t$, so lowering $\rho_t$ below $1-\epsilon$ yields no further gain. However, if an update has already pushed $\rho_t$ in the wrong direction (i.e., raising the probability of an action with $\hat{A}_t < 0$), then the unclipped term is selected, thereby retaining the gradient needed to correct the policy. The effect of this process is a pessimistic lower bound on policy improvement that discourages destructively large updates.

\subsection{Generalized Advantage Estimation} \label{app:ppo-gae}
    The advantage $\hat A_t$ in Equation~\ref{eq:ppo-clip} measures how much better action $a_t$ was than the expected behavior of the policy at $s_t$. PPO estimates advantage using Generalized Advantage Estimation (GAE), which uses a smoothing parameter $\lambda \in [0,1]$. Let $\delta_t$ denote the one-step temporal-difference (TD) residual computed with respect to the predicted value at rollout time:
    \begin{equation}
        \delta_t \;=\; r_t \;+\; \gamma\, V_{\theta_{\mathrm{old}}}(s_{t+1})\,(1 - d_{t+1}) \;-\; V_{\theta_{\mathrm{old}}}(s_t).
    \end{equation}
    The GAE advantage is then the exponentially weighted sum of $k$-step TD residuals,
    \begin{equation}\label{eq:gae}
        \hat{A}_t \;=\; \sum_{l=0}^{T-t-1} (\gamma\lambda)^{l}\,\delta_{t+l},
    \end{equation}
        or equivalently
    \begin{equation}\label{eq:gae-recursive}
        \hat{A}_t \;=\; \delta_t \;+\; \gamma\,\lambda\,(1 - d_{t+1})\,\hat{A}_{t+1},
        \qquad \hat{A}_T \;=\; 0,
    \end{equation}
    where $(1-d_{t+1})$ truncates the sum at episode terminations. If $\lambda = 0$, then $\hat{A}_t = \delta_t$ and, therefore, recovers only the one-step TD estimate (low variance, high bias); if $\lambda = 1$, then $\hat{A}_t$ approaches the full Monte-Carlo return (low bias, high variance). Intermediate values (typically $\lambda \in [0.9, 0.99]$) give a preferable bias-variance trade-off. Advantages are computed in a single backward recursion over the rollout, then normalized to zero mean and unit variance within each minibatch before being used in Equation~\ref{eq:ppo-clip}. 

\subsection{Clipped Value Loss} \label{app:ppo-value}
    Many implementations of PPO also clip the value loss similar to the clipping performed on the policy objective, bounding how much $V_\theta(s_t)$ can change from $V_{\theta_{\mathrm{old}}}(s_t)$ in a single update:
    \begin{equation}\label{eq:ppo-value}
        \mathcal{L}^{V}(\theta) \;=\; \tfrac{1}{2}\,\mathbb{E}_t\!\left[
        \max\!\Big( \big(V_\theta(s_t) - \hat{R}_t\big)^{2},\;
        \big(\tilde V_\theta(s_t) - \hat{R}_t\big)^{2} \Big) \right],
    \end{equation}
    with the clipped value prediction
    \begin{equation}
        \tilde V_\theta(s_t) \;=\; V_{\theta_{\mathrm{old}}}(s_t) \;+\; \mathrm{clip}\!\big(V_\theta(s_t) - V_{\theta_{\mathrm{old}}}(s_t),\; -\epsilon,\; \epsilon\big),
    \end{equation}
    so that the clipped prediction $\tilde V_\theta(s_t)$ lies within an $\epsilon$-ball of its rollout-time estimate. The max operator selects the prediction that yields the larger squared error. Therefore, when $V_\theta(s_t)$ moves outside the $\epsilon$-ball in a direction that would \textit{reduce} the loss, the clipped term is used; otherwise, the unclipped error is used. This makes $\mathcal{L}^{V}$ a pessimistic bound on $V_\theta(s_t)$, similar to the role of the min operator in $\mathcal{L}^{\mathrm{P}}$, and improves stability when reward magnitudes are large or the value function has not yet converged.

\subsection{Entropy Bonus}\label{app:ppo-entropy}
    Maximizing the surrogate alone tends to collapse the policy onto a near-deterministic behavior. Therefore, to preserve exploration, PPO regularizes the policy with the Shannon entropy of its action distribution,
    \begin{equation}\label{eq:ppo-entropy}
        \mathcal{L}^{E}(\theta) \;=\; \mathbb{E}_t\!\left[ H\!\big(\pi_\theta(\cdot \!\mid\! s_t)\big) \right],
    \end{equation}
    \begin{equation}
        H\!\big(\pi_\theta(\cdot \!\mid\! s_t)\big) \;=\; -\!\sum_{a \in \mathcal{A}}\! \pi_\theta(a \!\mid\! s_t)\,\log \pi_\theta(a \!\mid\! s_t),
    \end{equation}
    The coefficient $c_e$ determines the amount of exploration with large $c_e$ values keeping the action distribution flat, while small $c_e$ values allow high determinism as confidence increases. 

\subsection{Combined Objective} \label{app:ppo-loop}
    As presented in Equation~\ref{eq:ppo-loss}, the three terms of $\mathcal{L}^{\mathrm{P}}$, $\mathcal{L}^{V}$, $\mathcal{L}^{E}$ are combined into a single scalar loss to be minimized by stochastic gradient descent. Specifically, $\mathcal{L}^{\mathrm{P}}$ and $\mathcal{L}^{E}$ are the objectives to maximize, while $\mathcal{L}^{V}$ is a squared-error loss to be minimized. The pseudo code of PPO is presented in Algorithm~\ref{alg:ppo} and the hyperparameters used in this work are presented in Appendix~\ref{app:hyperparams}. 
    \begin{algorithm}[t]
        \caption{Proximal Policy Optimization (clipped).}
        \label{alg:ppo}
        \begin{algorithmic}[1]
            \Require initial parameters $\theta$, clip range $\epsilon$, discount $\gamma$, GAE parameter $\lambda$, coefficients $c_v, c_e$, epochs $K$, rollout length $T$, updates $U$
            \For{iteration $= 1, \ldots, U$}
                \State $\theta_{\mathrm{old}} \gets \theta$
                \State Roll out $\pi_{\theta_{\mathrm{old}}}$ for $T$ steps; 
                \State \quad store $(s_t, a_t, r_t, d_t, \log \pi_{\theta_{\mathrm{old}}}(a_t \!\mid\! s_t), V_{\theta_{\mathrm{old}}}(s_t))$
                \State Compute $\{\delta_t\}_{t=0}^{T-1}$ and $\{\hat{A}_t\}_{t=0}^{T-1}$ by Eq.~\ref{eq:gae-recursive}
                \State Compute value targets $\hat{R}_t \gets \hat{A}_t + V_{\theta_{\mathrm{old}}}(s_t)$
                \For{epoch $= 1, \ldots, K$}
                    \For{each minibatch $\mathcal{B} \subset \{0, \ldots, T-1\}$}
                        \State Normalize $\{\hat{A}_t\}_{t \in \mathcal{B}}$
                        \State Evaluate $\mathcal{L}^{\mathrm{P}}, \mathcal{L}^{V}, \mathcal{L}^{E}$ over $\mathcal{B}$ via Eqs.~\ref{eq:ppo-clip},~\ref{eq:ppo-value},~\ref{eq:ppo-entropy}
                        \State $\theta \gets \theta - \eta\, \nabla_\theta\, \mathcal{L}^{\mathrm{PPO}}(\theta)$ // gradient-norm clip
                    \EndFor
                \EndFor
            \EndFor
        \end{algorithmic}
      \end{algorithm}
    In practice the learning rate $\eta$ is linearly annealed over the course of training and the global gradient norm is clipped to a fixed value.

\section{Hyperparameters} \label{app:hyperparams}

\noindent\textbf{Problem-Space Constraints.} 
    We manage problem-space conflicts following the implementation of~\cite{cortellazzi2025intriguing}, which we requested from the authors. 
    Therefore, prior to selection, action masking removes: (i) gadgets whose Java classes conflict with already-transplanted gadgets, and (ii) gadgets that would exceed the per-gadget budgets of $3$ permissions and $60$ classes (as set in the provided implementation).

\subsection{\replicant} \label{app:hyper-rep} 
    We train \replicant with PPO~\cite{schulman2017proximal} using the default hyperparameters of CleanRL~\cite{huang2022cleanrl}: learning rate $\alpha=2.5{\times}10^{-4}$ (annealed), $\gamma{=}0.99$, GAE $\lambda{=}0.95$, clip $\varepsilon{=}0.2$, entropy coefficient $0.01$, value coefficient $0.5$ with clipped value loss, max gradient norm $0.5$, and normalized advantages. 
    Each \replicant agent is trained for $50$ updates, with each update using $500$ episodes per batch, $250$ steps per minibatch, and $3$ epochs over all minibatches.
    Each episode runs for a maximum horizon of $H=100$ steps, with a forced \textsc{Submit} every $p=10$ steps. 
    The architecture of \replicant is based on DRMD~\cite{mcfadden2026drmd}, given its state-of-the-art detection performance in the domain, but extended to learn the hierarchical evasion policy.
    Specifically, \replicant uses a shared two-layer MLP encoder ($512$ neurons, LeakyReLU, dropout) feeding three heads, each adding one further $512$-neuron layer: a \textit{query} head over $\{\textsc{Submit},\textsc{Modify}\}$, a \textit{modify} head over the capability set $C$, and a \textit{critic} head estimating $V(s_t)$. 
    During training, the agent samples from the action distribution to allow for effective action space exploration, whereas during evaluation it selects the highest-probability action to exploit the knowledge learned in training.   

\subsection{Classifiers} \label{app:hyper-clf}

 \noindent\textbf{RandomForest}~\cite{MaMaDroid}. Scikit-learn implementation with $101$ estimators, max depth $64$, and parallel fitting ($n\_jobs{=}-2$).

  \noindent\textbf{LinearSVC}~\cite{Drebin}. Scikit-learn LinearSVC with max iterations $50{,}000$ and default regularization.

  \noindent\textbf{SecSVM}~\cite{demontis2017yes}. trained via full-batch GPU sub-gradient descent with $C{=}1.0$, weight bounds $[-0.5, 0.5]$, learning rate $\eta{=}0.5$, convergence threshold $\epsilon{=}10^{-4}$, and max $10{,}000$ iterations.

  \noindent\textbf{DeepDrebin}~\cite{deepDrebin}. Two-hidden-layer MLP ($200{\times}200$, ReLU) trained with SGD at $\text{lr}{=}0.05$ for $10$ epochs (batch size $512$), repeated over $5$ training loops with validation-based checkpoint
  selection.

  \noindent\textbf{DRMD}~\cite{mcfadden2026drmd}. PPO-based \ac{DRL} classifier with a $3{\times}512$-unit MLP (LeakyReLU, dropout $0.5$), trained for $5$ epochs with learning rate $2.5{\times}10^{-4}$ (annealed), $\gamma{=}0.99$, GAE $\lambda{=}0.95$, clip $\varepsilon{=}0.2$, entropy coefficient $0.01$, minibatch size $250$, and $1$ PPO update epoch per training epoch.

  \noindent\textbf{HCC}~\cite{chen2023continuous}. Contrastive learning classifier with a $4$-layer encoder ($512{\to}384{\to}256{\to}128$, ReLU) and a $3$-layer MLP head ($128{\to}100{\to}100{\to}2$, ReLU, dropout $0.2$). Trained for $40$ epochs (batch size $1{,}024$) with contrastive loss weight $\lambda{=}100$ and margin $10$.

  \noindent\textbf{RAMDA}~\cite{ramda}. Two-phase architecture: a VAE ($n_{\text{features}}{\to}600{\to}600{\to}80$ latent, dropout $0.1$) followed by a $4$-layer detection MLP ($80{\to}40{\to}40{\to}40{\to}2$, dropout $0.1$). Both trained for $50$ epochs with a threshold $v{=}30$.

\subsection{Baselines} \label{app:hyper-base}
    For fair comparison and to maintain realizability, all baselines use the same problem-space constraints as \replicant.

    \noindent\textbf{APG}~\cite{pierazzi2020intriguing,cortellazzi2025intriguing}. Greedily selects capabilities ordered by their contribution to the decision function of the classifier. For linear classifiers (LinearSVC, SecSVM), contribution is computed via weight-based projection; for neural classifiers, a score-based approximation is used.

    \noindent\textbf{ADZ}~\cite{he2023efficient}. Operates over a perturbation space of $662$ actions ($143$ manifest features and $519$ code features) organized in a hierarchical selection tree. At each step, a perturbation is sampled from the tree and applied, then the modified sample is submitted to the classifier and the tree is updated based on the change in the confidence of the classifier. As shown in Appendix~\ref{tab:adz-compare}, the ADZ method does not scale to larger capability sets, therefore, ADZ is the only exception to the shared capability set across attackers. We instead reconstruct a $662$-action space from our capability set following the subset of manifest and code perturbations specified in the original work.

    \noindent\textbf{EvadeDroid}~\cite{bostani2024evadedroid}. At each step, samples and applies an untried capabilities at random, and queries the target classifier. If the sampled capability does not decrease the classifier confidence then it is removed from the malware. Every iteration queries the target classifier once. Results are averaged over $5$ independent trials per seed.

    \noindent\textbf{Rnd}~\cite{mcfadden2026sok}. At each step, randomly selects a query action (MODIFY/SUBMIT) and a capability, uniformly at random, from the valid (masked) action set. Results are averaged over $5$ independent trials per seed.

\subsection{RL-Based Attacks} \label{app:hyper-rl}

    \noindent\textbf{MEME}~\cite{rigaki2023power}. A PPO attack for model-extraction and evasion; we collapse its extraction stage and instead train the agent directly against the shared surrogate using the Optuna-tuned hyperparameters from the paper (lr$=1.38{\times}10^{-3}$, $\gamma{=}0.854$, max-grad-norm$=0.428$, $10$ epochs) with SB3 (Stable Baselines 3) defaults otherwise ($n_{\text{steps}}{=}128$, minibatch $64$, GAE $\lambda{=}0.95$, clip $0.2$, entropy $0$, value $0.5$), over the paper's fixed training budget of $6{,}144$ environment steps. The reward function of MEME is $+10$ on evasion and otherwise the step-wise reduction in the malicious score of the surrogate.

    \noindent\textbf{MAB}~\cite{song2022mab}. Thompson-sampling multi-armed bandit in which each capability is an arm with a Beta$(\alpha,\beta)$ posterior initialized at $\alpha{=}\beta{=}1$. At each step it draws a value from every arm and applies the highest-drawn valid capability; a leave-one-out minimization pass then prunes redundant capabilities and credits the posteriors from the outcome ($\alpha{+}1$ for capabilities whose removal breaks evasion, $\beta{+}1$ for redundant ones), reinforcing the arms useful for evasion. Results are averaged over $5$ trials per seed.

    \noindent\textbf{MERLIN}~\cite{quertier2022merlin}. Deep Q Network (DQN) with a two-layer $128$-unit MLP (ReLU), Adam lr$=2.5{\times}10^{-4}$, $\gamma{=}0.99$, uniform replay of $10$k, batch $32$, hard target-update every $500$ steps, $1{,}000$ training episodes, and $\varepsilon$ annealed to $0.05$ over the first half of training, rewarded $+10$ on evasion and otherwise by the step-wise reduction in the confidence score of the surrogate.

    \noindent\textbf{AIMED-RL}~\cite{labaca2021aimed}. Distributional Double-DQN with NoisyNet exploration ($\sigma_0{=}0.5$) and proportional prioritized replay ($\alpha{=}0.6$, $\beta{:}0.4{\to}1$); $\gamma{=}0.95$, Adam(lr$=10^{-3}$, $\epsilon{=}10^{-2}$), batch $32$, target-update every $100$ steps, and a $1{,}000$ episodes training budget. Its reward function is comprised of an evasion bonus ($+10$), a distance reward in terms of capabilities applied, a similarity reward in terms of the actual changes to the file, and a penalty for repeated capabilities.

    \noindent\textbf{PSP-Mal}~\cite{zhan2023psp}. Dueling Double-DQN with soft target updates ($\tau{=}0.005$), Adam lr$=3{\times}10^{-4}$, $\gamma{=}0.99$, batch $256$, replay of $500$k, $\varepsilon$ annealed $1.0{\to}0.1$, and a training budget of $1{,}000$ episodes. Its reward function is $+10$ on evasion and otherwise the reduction of the confidence score from the original sample. Its prioritized replay adds a Shapley-style prior, which we compute on the surrogate to maintain parity between attackers. 

\section{Scale of Experiments}\label{sec:app-scale}
    Across all experiments for the main results of the paper we conduced $379,680$ attack evaluations, we further breakdown this total by section.
    For Sections~\ref{sec:eval-query}-\ref{sec:eval-transfer}, the ten attacks are evaluated across $1{,}764$ surrogate/target combinations ($7{\times}7$ classifiers, $3{\times}3$ feature spaces, and $2{\times}2$ training periods), two dataset variants, and ten seeds, totaling $352{,}800$ attack evaluations.
    For Sections~\ref{sec:eval-at}-\ref{sec:eval-capdrift}, $10{,}080$ attack evaluations and $560$ hardened classifiers ($8{,}400$ \ac{AT} rounds) were produced across the seven classifiers, ten seeds, two training periods, two pipelines, two \ac{AT} capability sets, and six attacker capability sets. 
    Lastly, Section~\ref{sec:eval-temporal} tracks monthly performance of five detection pipelines across seven classifiers, two training periods, and ten seeds, leading to $16{,}800$ attack evaluations across the $2022$--$2024$ period.
    To our knowledge this is the largest evaluation of malware evasion to date, delineating the performance differences of attack \textit{strategies} rather than that of capability set or threat model.

\section{Extended Evaluations}\label{app:eval-robust}
\begin{figure}[H]
        \centering
        \includegraphics[width=\linewidth]{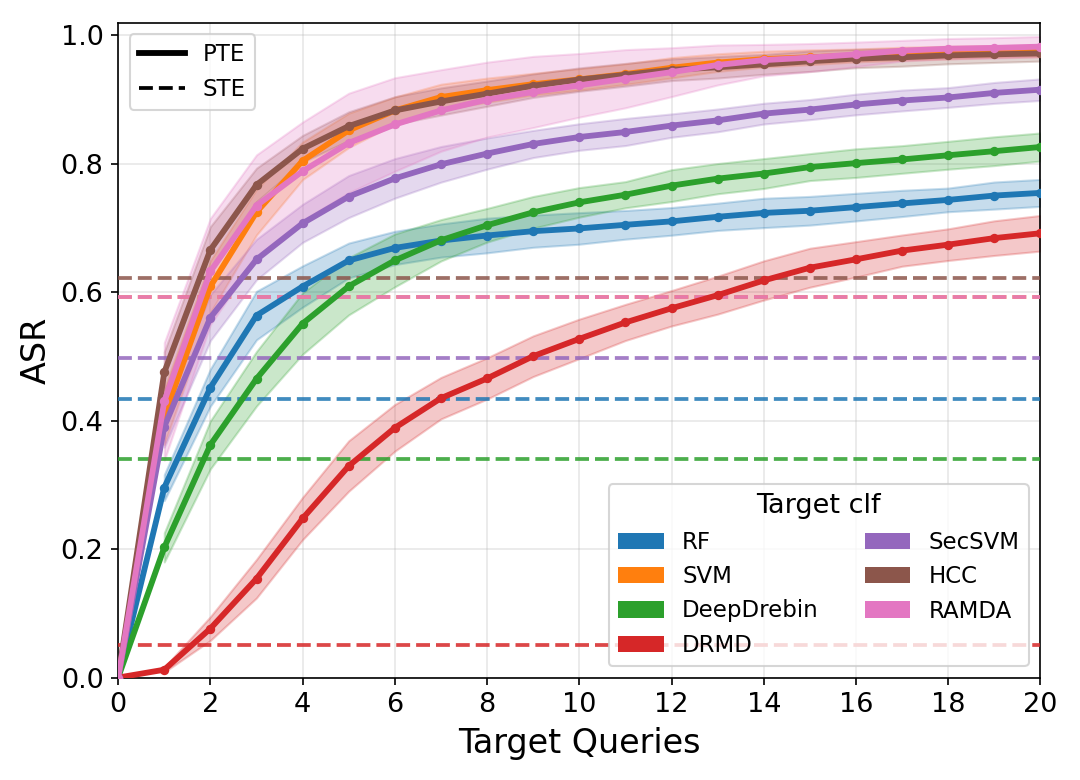}
        \caption{\textbf{Architecture gap}, per target classifier: Solid curves plot the \ac{ASR} of \ac{PTE} over target queries and the dashed horizontals lines demonstrate the \ac{ASR} of \ac{STE}. Aggregated as the \textit{architecture gap} row of Table~\ref{tab:transfer} and the green curve of Figure~\ref{fig:transfer}.}
        \label{fig:transfer-arch}
    \end{figure}

\begin{figure}[H]
        \centering
        \includegraphics[width=\linewidth]{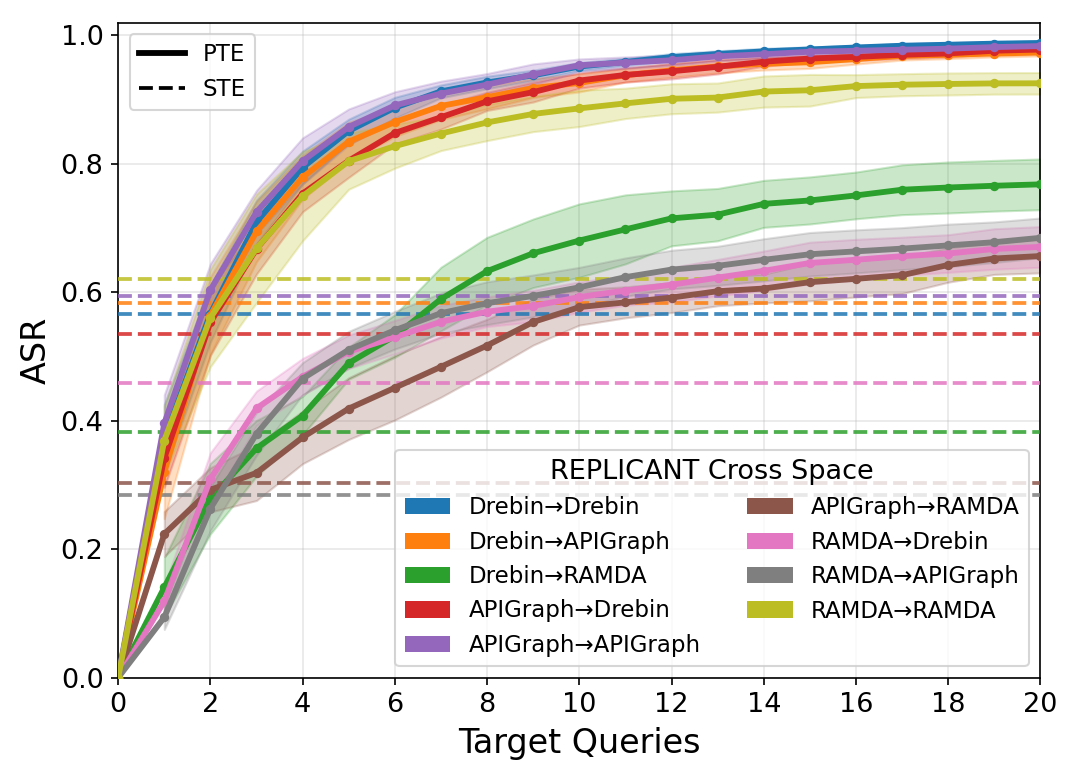}
        \caption{\textbf{Representation gap}, per feature-space pair: Solid curves plot the \ac{ASR} of \ac{PTE} over target queries and the dashed horizontals lines demonstrate the \ac{ASR} of \ac{STE}. Aggregated as the \textit{representation gap} row of Table~\ref{tab:transfer} and the orange curve of Figure~\ref{fig:transfer}.}
        \label{fig:transfer-repr}
    \end{figure}

\begin{figure}[H]
        \centering
        \includegraphics[width=\linewidth]{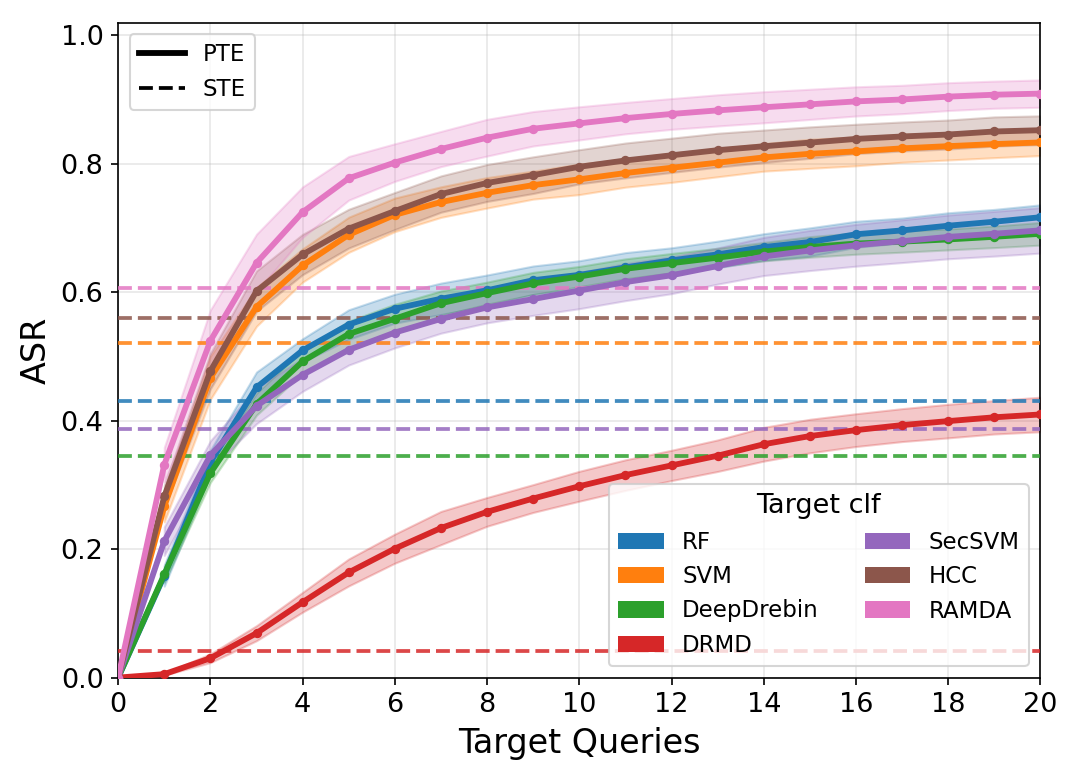}
        \caption{\textbf{Combined gap}, per target classifier: Solid curves plot the \ac{ASR} of \ac{PTE} over target queries and the dashed horizontals lines demonstrate the \ac{ASR} of \ac{STE}. Aggregated as the \textit{combined gap} row of Table~\ref{tab:transfer} and the red curve of Figure~\ref{fig:transfer}.}
        \label{fig:transfer-combined}
    \end{figure}

    The following tables and figures support the evaluations described in Section~\ref{sec:eval-query}: the effect of the VirusTotal labeling threshold on FNR (Table~\ref{tab:fnr}); the attack results on Hypercube$_{VTT4}$ (Table~\ref{tab:baselines-vtt4}); the attack results for the different time-period splits of the Hypercube (VTT2) dataset (Table~\ref{tab:idood}); and, the per-classifier and per-space-pair breakdowns of the transferability curves that Figure~\ref{fig:transfer} aggregates (Figures~\ref{fig:transfer-arch}-\ref{fig:transfer-combined}).

        \begin{table}[h]
            \centering
            \caption{\textbf{FNRs between VTTs}, target classifier false-negative rate (FNR, \%) on malware by period. Reported using $2021b$ targets for full dataset coverage. $2021a$ is the surrogate held-out set; $2022$--$2024$ are the temporally split test years. FNRs are averaged over the seven target classifiers, three feature spaces, and ten seeds; the final column averages across the three periods. As Hypercube$_{VTT4}$ contains no $2024$ malware, its $2024$ entry is omitted and its mean is taken over $2021$--$2023$.}
            \vspace{3pt}
            \label{tab:fnr}
             \adjustbox{max width=\columnwidth}{%
            \begin{tabular}{lccccc}
              \toprule
              Dataset & $2021a$ & $2022$ & $2023$ & $2024$ & Mean \\
              \midrule
              VTT2 & $35.6_{\pm 0.4}$ & $63.8_{\pm 0.6}$ & $86.6_{\pm 0.5}$ & $95.0_{\pm 0.4}$ & $70.2_{\pm 0.3}$ \\
              VTT4 & $15.6_{\pm 0.2}$ & $46.9_{\pm 0.6}$ & $66.2_{\pm 0.5}$ & $-$ & $42.9_{\pm 0.3}$ \\
              \bottomrule
            \end{tabular}
            }
          \end{table}

        \begin{table}[H]
            \centering
            \caption{\textbf{VTT4 Results}, Attacker comparison on Hypercube$_{VTT4}$ for both the \textit{matched} and \textit{all} settings, reporting mean \ac{ASR} and AvgQ for both, with per target breakdown for matched.}
            \vspace{3pt}
            \label{tab:baselines-vtt4}
            \adjustbox{max width=\columnwidth}{%
            \begin{tabular}{lccccc}
                \toprule
                \textbf{Matched} & \textbf{\replicant} & \textbf{APG} & \textbf{ADZ} & \textbf{EvD} & \textbf{Rnd} \\
                \cmidrule(lr){1-6}
                  SVM        & $\mathbf{100.0}^{***}_{\pm 0.0}$ & $\mathbf{100.0}^{***}_{\pm 0.0}$ & $88.2_{\pm 0.9}$ & $\textit{96.6}_{\pm 0.2}$ & $91.0_{\pm 1.1}$ \\
                  SecSVM     & $\mathbf{100.0}^{***}_{\pm 0.0}$ & $\mathbf{100.0}^{***}_{\pm 0.0}$ & $71.4_{\pm 3.4}$ & $\textit{89.2}_{\pm 0.8}$ & $74.5_{\pm 2.6}$ \\
                  RF         & $\mathbf{93.8}^{***}_{\pm 1.6}$ & $91.0_{\pm 1.6}$ & $90.0_{\pm 1.6}$ & $91.1_{\pm 1.6}$ & $\textit{91.1}_{\pm 1.6}$ \\
                  DeepDrebin & $\mathbf{100.0}^{***}_{\pm 0.0}$ & $\mathbf{100.0}^{***}_{\pm 0.0}$ & $86.9_{\pm 1.9}$ & $\textit{91.9}_{\pm 1.6}$ & $88.3_{\pm 1.4}$ \\
                  RAMDA      & $\mathbf{99.8}^{***}_{\pm 0.3}$ & $83.7_{\pm 3.6}$ & $84.3_{\pm 2.8}$ & $77.9_{\pm 2.9}$ & $\textit{86.4}_{\pm 1.4}$ \\
                  HCC        & $\mathbf{99.9}^{**}_{\pm 0.1}$ & $\textit{99.3}_{\pm 0.7}$ & $90.4_{\pm 1.4}$ & $92.5_{\pm 1.6}$ & $92.0_{\pm 1.4}$ \\
                  DRMD       & $\mathbf{100.0}^{***}_{\pm 0.0}$ & $59.7_{\pm 3.0}$ & $60.9_{\pm 2.6}$ & $\textit{74.9}_{\pm 1.7}$ & $57.0_{\pm 1.4}$ \\
                \cmidrule(lr){1-6}
                    ASR & $\mathbf{99.0}^{***}_{\pm 0.2}$ & $\textit{90.5}_{\pm 0.9}$ & $81.8_{\pm 1.6}$ & $87.7_{\pm 1.0}$ & $82.9_{\pm 0.6}$ \\
                    AvgQ     & $\phantom{0}\mathbf{2.6}^{***}_{\pm 0.1}$ & $\phantom{0}\mathbf{3.8}^{***}_{\pm 0.2}$ & $\phantom{0}7.5_{\pm 0.3}$ & $\phantom{0}\textit{6.5}_{\pm 0.2}$ & $\phantom{0}6.6_{\pm 0.2}$ \\
                \midrule
                \textbf{All Settings} & \textbf{\replicant} & \textbf{APG} & \textbf{ADZ} & \textbf{EvD} & \textbf{Rnd} \\
                \cmidrule(lr){1-6}
                    ASR & $\mathbf{94.1}^{***}_{\pm 0.9}$ & $\textit{83.8}_{\pm 1.1}$ & $82.0_{\pm 1.5}$ & $78.2_{\pm 1.2}$ & $81.8_{\pm 1.0}$ \\
                    AvgQ     & $\phantom{0}\mathbf{5.1}^{***}_{\pm 0.3}$ & $\phantom{0}\textit{6.0}_{\pm 0.3}$ & $\phantom{0}7.4_{\pm 0.3}$ & $\phantom{0}8.1_{\pm 0.3}$ & $\phantom{0}6.6_{\pm 0.3}$ \\

                \bottomrule
            \end{tabular}}
        \end{table}

        \begin{table}[H]
            \centering
            \caption{\textbf{Test Splits Performance}, \ac{ASR} for the attacks over all settings by test period \textit{relative} to the training year of the target, averaged over the $2021b$- and $2022b$-trained targets. $+0$ is the surrogate split of the training year ($2021a$ for a $2021b$ target, $2022a$ for a $2022b$ target); $+1$ is the main test setting used in Sections~\ref{sec:eval-query}-\ref{sec:eval-transfer}; $+2$/$+3$ test further into the future. $+3$ has only the $2021b$-trained targets as dataset does not contain $2025$ samples.}
            \vspace{3pt}
            \label{tab:idood}
            \adjustbox{max width=\columnwidth}{%
            \begin{tabular}{lccccc}
              \toprule
              Test period & \textbf{\replicant} & \textbf{APG} & \textbf{ADZ} & \textbf{EvD} & \textbf{Rnd} \\
              \midrule
              $+0$ & ${\mathbf{70.0}}^{***}_{\pm 1.3}$ & $\textit{53.4}_{\pm 0.7}$ & $49.7_{\pm 1.5}$ & $43.8_{\pm 0.8}$ & $46.2_{\pm 1.0}$ \\
              $+1$ & ${\mathbf{78.7}}^{***}_{\pm 1.2}$ & $\textit{65.0}_{\pm 0.9}$ & $61.0_{\pm 1.2}$ & $56.4_{\pm 0.9}$ & $58.9_{\pm 0.9}$ \\
              $+2$ & ${\mathbf{89.4}}^{***}_{\pm 1.3}$ & $\textit{82.8}_{\pm 0.9}$ & $77.5_{\pm 1.4}$ & $75.7_{\pm 1.4}$ & $77.1_{\pm 1.3}$ \\
              $+3$ & ${\mathbf{95.4}}^{***}_{\pm 1.2}$ & $\textit{91.9}_{\pm 1.2}$ & $86.8_{\pm 1.5}$ & $86.8_{\pm 1.6}$ & $87.6_{\pm 1.2}$ \\
              \midrule
              All & ${\mathbf{73.1}}^{***}_{\pm 1.2}$ & $\textit{57.2}_{\pm 0.7}$ & $53.5_{\pm 1.3}$ & $48.0_{\pm 0.7}$ & $50.5_{\pm 0.9}$ \\
              \bottomrule
            \end{tabular}
            }
          \end{table}

    \begin{figure}[H]
        \centering
        \includegraphics[width=\linewidth]{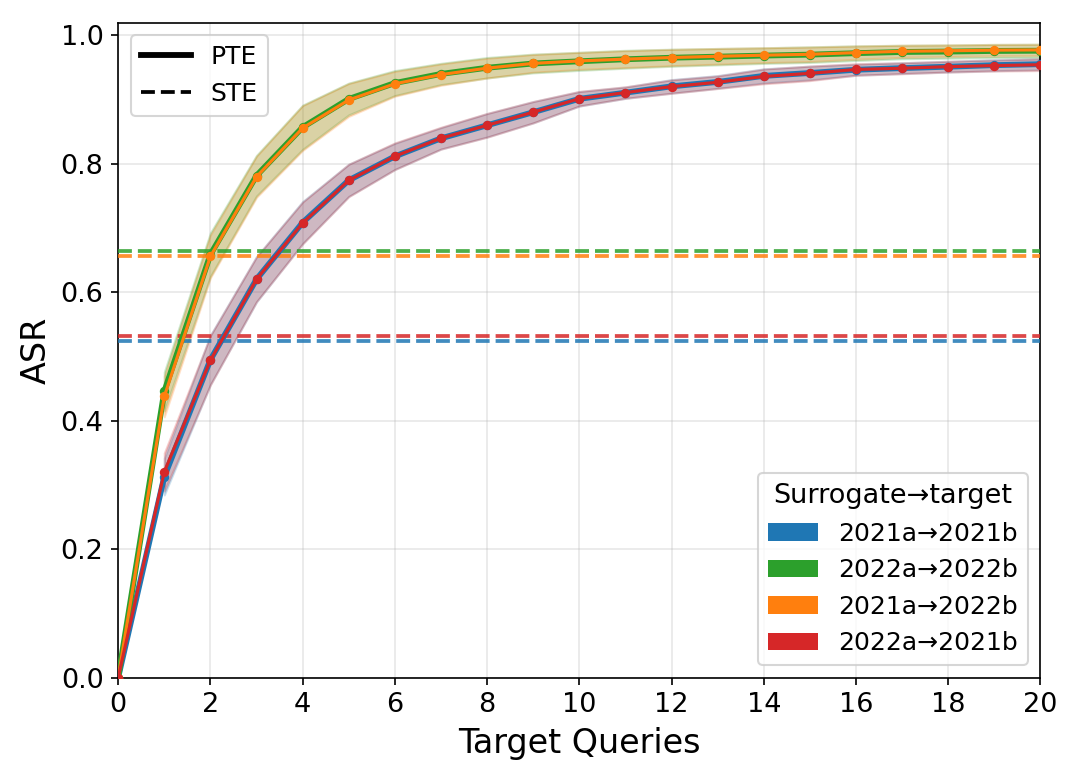}
        \caption{\textbf{Temporal gap}, per training-period pair: Solid curves plot the \ac{ASR} of \ac{PTE} over target queries and the dashed horizontals lines demonstrate the \ac{ASR} of \ac{STE}. Aggregated as the \textit{temporal gap} row of Table~\ref{tab:transfer} and the purple curve of Figure~\ref{fig:transfer}. The cross-period curves follow the matched-period curves with the same target training period, showing both the effect of temporal luck and that the temporal gap is negligible.}
        \label{fig:transfer-temporal}
    \end{figure}

    \section{RL Design Ablation} \label{app:rl-ab}
\begin{figure}[H]
        \centering
        \includegraphics[width=\linewidth]{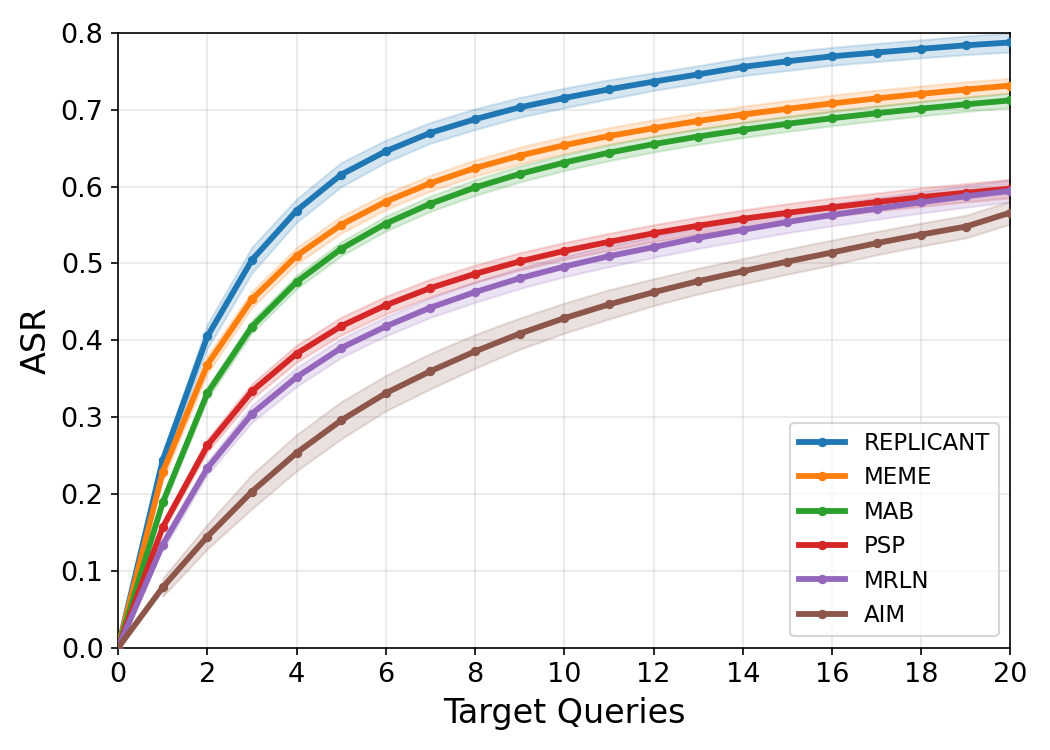}
        \caption{\textbf{RL Design Curves.} Attack efficiency across all $1{,}764$ surrogate/target combinations and ten seeds. \replicant outperforms every other RL agent design with the highest evasion rate per query.}
        \label{fig:allattackwinodws}
    \end{figure}

    To isolate the gains provided by the architecture of our agent and its reward function over prior RL-based work, we compare against five alternative state-of-the-art RL agent designs for malware evasion: MEME~\cite{rigaki2023power}, MAB~\cite{song2022mab}, MRLN~\cite{quertier2022merlin}, AIM~\cite{labaca2021aimed}, PSP~\cite{zhan2023psp}.
    As each of these five RL-based attacks were originally proposed for Windows PE malware evasion, we adapted them to Android by modifying them to use our environment, threat model, and capability set. 
    Results for this comparison can be seen in Table~\ref{tab:baselines-windows} and Figure~\ref{fig:allattackwinodws}.

    \begin{table}[H]
            \centering
            \caption{\textbf{RL Design Comparison.} Comparison between \replicant and prior RL-based attacks designs for Windows PE malware adapted to our environment, threat model, and shared capability set. Results are evaluated and reported following the methodology of Table~\ref{tab:baselines}. \replicant remains the strongest attacker with statistically significant gains across all settings.}
            \vspace{3pt}
            \label{tab:baselines-windows}
            \adjustbox{max width=\columnwidth}{%
            \begin{tabular}{lcccccc}
                \toprule
                \textbf{Matched} & \textbf{\replicant} & \textbf{MEME} & \textbf{MAB} & \textbf{PSP} & \textbf{MRLN} & \textbf{AIM} \\
                \cmidrule(lr){1-7}
                SVM        & $\mathbf{99.1}^{**}_{\pm 1.5}$ & $\textit{98.6}_{\pm 1.5}$ & $98.0_{\pm 2.3}$ & $77.3_{\pm 5.2}$ & $71.0_{\pm 6.6}$ & $72.0_{\pm 9.4}$ \\
                SecSVM    & $\mathbf{98.2}^{***}_{\pm 1.0}$ & $\textit{95.1}_{\pm 1.2}$ & $94.1_{\pm 1.0}$ & $59.6_{\pm 4.7}$ & $58.7_{\pm 4.3}$ & $45.7_{\pm 8.1}$ \\
                RF         & $\mathbf{93.3}^{***}_{\pm 1.7}$ & $74.9_{\pm 2.6}$ & $\textit{76.2}_{\pm 2.9}$ & $67.0_{\pm 3.0}$ & $62.1_{\pm 2.8}$ & $65.7_{\pm 3.8}$ \\
                DeepDrebin & $\mathbf{96.5}^{***}_{\pm 2.1}$ & $\textit{84.7}_{\pm 3.5}$ & $72.4_{\pm 2.9}$ & $59.4_{\pm 2.5}$ & $51.1_{\pm 4.2}$ & $49.3_{\pm 5.6}$ \\
                RAMDA      & $\mathbf{99.9}^{***}_{\pm 0.1}$ & $\textit{95.8}_{\pm 4.9}$ & $95.5_{\pm 5.6}$ & $83.4_{\pm 3.3}$ & $76.6_{\pm 9.3}$ & $80.0_{\pm 5.6}$ \\
                HCC        & $\mathbf{98.7}^{***}_{\pm 0.9}$ & $\textit{96.8}_{\pm 1.5}$ & $96.3_{\pm 1.6}$ & $80.0_{\pm 3.3}$ & $78.9_{\pm 4.0}$ & $78.8_{\pm 4.3}$ \\
                DRMD       & $\mathbf{90.2}^{***}_{\pm 2.7}$ & $\textit{83.1}_{\pm 4.4}$ & $66.7_{\pm 2.5}$ & $20.6_{\pm 2.4}$ & $31.7_{\pm 7.5}$ & $17.1_{\pm 4.9}$ \\
                \cmidrule(lr){1-7}
                \ac{ASR}   & $\mathbf{96.6}^{***}_{\pm 0.7}$ & $\textit{89.8}_{\pm 1.1}$ & $85.6_{\pm 1.6}$ & $63.9_{\pm 1.8}$ & $61.5_{\pm 2.0}$ & $58.4_{\pm 2.2}$ \\
                AvgQ       & $\phantom{0}\mathbf{3.6}^{***}_{\pm 0.3}$ & $\phantom{0}\textit{5.1}_{\pm 0.3}$ & $\phantom{0}6.0_{\pm 0.3}$ & $10.2_{\pm 0.3}$ & $11.5_{\pm 0.3}$ & $13.8_{\pm 0.3}$ \\
                \midrule
                \textbf{All Settings} & \textbf{\replicant} & \textbf{MEME} & \textbf{MAB} & \textbf{PSP} & \textbf{MRLN} & \textbf{AIM} \\
                \cmidrule(lr){1-7}
                \ac{ASR}   & $\mathbf{78.8}^{***}_{\pm 1.2}$ & $\textit{73.2}_{\pm 0.9}$ & $71.2_{\pm 0.9}$ & $59.7_{\pm 1.2}$ & $59.5_{\pm 1.4}$ & $56.6_{\pm 1.5}$ \\
                AvgQ       & $\phantom{0}\mathbf{7.4}^{***}_{\pm 0.2}$ & $\phantom{0}\textit{8.5}_{\pm 0.2}$ & $\phantom{0}9.0_{\pm 0.2}$ & $11.0_{\pm 0.2}$ & $11.3_{\pm 0.2}$ & $12.6_{\pm 0.3}$ \\
                \bottomrule
            \end{tabular}}
        \end{table}

    \section{Motivating Ablations}\label{app:ablations}
    This appendix motivates three design decisions: the scaling of training budget in \atreplicant (Table~\ref{tab:at-scaling}); the reduced ADZ capability set (Table~\ref{tab:adz-compare}), and the effect of surrogate/test overlap in the $2022a{\rightarrow}2021b$ setting (Table~\ref{tab:overlap}).

    \begin{table}[H]
        \centering
        \caption{\textbf{Ablation: ADZ Capability Set}, the mean \ac{ASR} performance of ADZ decreases when it is given the full capability set, demonstrating that it does not scale and motivating our use of the reduced $662$-set following the constraints of the original work~\cite{he2023efficient}. Although reduced when using the smaller $662$-set, \replicant still remains significantly stronger than ADZ. Experiments conducted using: the $2021a{\rightarrow}2021b$ training period, Drebin feature space, five seeds, and seven classifiers; results reported as mean \ac{ASR} for matched settings.}
        \vspace{3pt}
        \label{tab:adz-compare}
        \adjustbox{max width=\columnwidth}{%
      \begin{tabular}{lcc}
            \toprule
            \textbf{Strategy} & \textbf{ADZ $662$-set} & \textbf{Full Capability Set} \\
            \midrule
            ADZ        & $51.8_{\pm 1.5}$ & $34.7_{\pm 3.9}$ \\
            \replicant & $76.6_{\pm 4.5}$ & $95.5_{\pm 0.9}$ \\
            \bottomrule
        \end{tabular}}
    \end{table}

    \begin{table}[H]
        \centering
        \caption{\textbf{Ablation: Surrogate Overlap}, evaluation of the impact that the inclusion of $2022a$ data has on the mean \ac{ASR} for the $2022a{\to}2021b$ setting. \textit{Full} uses the entire $2022$ test set; \textit{excl. overlap} removes the test samples present in the $2022a$ surrogate training split. The change $\Delta$ is below $0.2$ point in both settings. Experiments conducted using: the $2022a{\to}2021b$ training periods, Drebin feature space, five seeds, and seven classifiers.}
        \vspace{3pt}
        \label{tab:overlap}
        \adjustbox{max width=\columnwidth}{%
      \begin{tabular}{lccc}
            \toprule
            \textbf{Setting} & \multicolumn{1}{c}{\textbf{ASR} \textbf{(full)}} & \multicolumn{1}{c}{\textbf{ASR} \textbf{(excl. overlap)}} & \multicolumn{1}{c}{$\Delta$} \\
            \midrule
            Matched & $98.16_{\pm 0.6}$ & $98.11_{\pm 0.6}$ & $+0.05$ \\
            All     & $86.27_{\pm 0.9}$ & $86.11_{\pm 0.6}$ & $+0.16$ \\
            \bottomrule
        \end{tabular}}
    \end{table}

    \begin{table}[H]
        \centering
        \caption{\textbf{Ablation: \ac{AT} Scaling}, unique AEs collected per \atreplicant round with and without doubling training length every five rounds. The two schedules are near-identical over the first five rounds; however, the constant budget collapses monotonically as the target hardens, whereas the increasing budget continues to find AEs throughout the 15 rounds. Experiments conducted using: the $2021a{\rightarrow}2021b$ training period, Drebin feature space, five seeds, and seven classifiers.}
        \vspace{3pt}
        \label{tab:at-scaling}
        \adjustbox{max width=\textwidth}{%
        \begin{tabular}{lccccc}
          \toprule
          \textbf{Round} & $1$ & $2$ & $3$ & $4$ & $5$ \\
          Increasing & $16209$ & $11398$ & $6782$ & $4060$ & $1717$ \\
          Constant & $15776$ & $11461$ & $7329$ & $4077$ & $1930$ \\
          \midrule
           & $6$ & $7$ & $8$ & $9$ & $10$ \\
          Increasing & $\mathbf{1998}$ & $1635$ & $913$ & $558$ & $541$ \\
          Constant & $865$ & $352$ & $211$ & $159$ & $134$ \\
          \midrule
           & $11$ & $12$ & $13$ & $14$ & $15$ \\
          Increasing & $\mathbf{3444}$ & $1464$ & $1131$ & $384$ & $221$ \\
          Constant & $88$ & $71$ & $68$ & $61$ & $82$ \\
          \bottomrule
        \end{tabular}}
    \end{table}

\end{document}